\documentclass[10pt,twocolumn,letterpaper]{article}

\usepackage[pagenumbers]{arxiv}

\newcommand{\bx}{\mathbf{x}}
\newcommand{\bh}{\mathbf{h}}
\newcommand{\bs}{\mathbf{s}}
\newcommand{\bb}{\mathbf{b}}
\newcommand{\bW}{\mathbf{W}}
\newcommand{\btheta}{\mathbf{\theta}}
\newcommand{\bpsi}{\mathbf{\psi}}

\newcommand{\method}{\textsc{DiffLens}\xspace}

\newcommand{\nocomment}{} 

\ifdefined\nocomment

\definecolor{arxivblue}{rgb}{0.21,0.49,0.74}
\usepackage[pagebackref,breaklinks,colorlinks,allcolors=arxivblue]{hyperref}
\usepackage[linesnumbered,ruled,vlined,english,onelanguage]{algorithm2e} 
\usepackage{float}
\usepackage{algpseudocode}
\usepackage{multirow}
\usepackage{wrapfig}
\usepackage{svg}
\usepackage{verbatim}
\usepackage{amssymb}
\usepackage{pifont}
\usepackage{bbding}
\usepackage{graphicx}
\usepackage{amsmath}
\usepackage{amssymb}
\usepackage{multirow}
\usepackage{booktabs}
\usepackage{pifont} 
\usepackage{xcolor}
\usepackage{tikz}

\usetikzlibrary{arrows.meta, positioning}

\newcommand\tikzmark[1]{\tikz[overlay,remember picture] \node (#1) {};}

\title{Dissecting and Mitigating Diffusion Bias via Mechanistic Interpretability}
\author{
  \quad \textbf{Yingdong Shi}$^{1\,*}$ 
  \quad \textbf{Changming Li}$^{1\,*}$ 
  \quad \textbf{Yifan Wang}$^{2}$ 
  \quad \textbf{Yongxiang Zhao}$^{1}$ \\
  \quad \textbf{Anqi Pang}$^{3}$ 
  \quad \textbf{Sibei Yang}$^{1}$ 
  \quad \textbf{Jingyi Yu}$^{1}$ 
  \quad\textbf{Kan Ren}$^{1\dagger}$\\ \\
  \textsuperscript{1}ShanghaiTech University
  \quad \textsuperscript{2}Stony Brook University
  \quad\textsuperscript{3}Tencent PCG  \\
  \\
{\tt\small\{shiyd2023,lichm2024,renkan\}@shanghaitech.edu.cn}
}

\begin{document}
\maketitle
\renewcommand{\thefootnote}{\fnsymbol{footnote}}
\footnotetext[1]{Equal contribution. \textsuperscript{$\dagger$}Corresponding author.}
\renewcommand*{\thefootnote}{\arabic{footnote}}
\begin{abstract}
Diffusion models have demonstrated impressive capabilities in synthesizing diverse content. 
However, despite their high-quality outputs, these models often perpetuate social biases, including those related to gender and race.
These biases can potentially contribute to harmful real-world consequences, reinforcing stereotypes and exacerbating inequalities in various social contexts.
While existing research on diffusion bias mitigation has predominantly focused on guiding content generation, it often neglects the intrinsic mechanisms within diffusion models that causally drive biased outputs. 
In this paper, we investigate the internal processes of diffusion models, identifying specific decision-making mechanisms, termed bias features, embedded within the model architecture.
By directly manipulating these features, our method precisely isolates and adjusts the elements responsible for bias generation, permitting granular control over the bias levels in the generated content.
Through experiments on both unconditional and conditional diffusion models across various social bias attributes, we demonstrate our method's efficacy in managing generation distribution while preserving image quality.
We also dissect the discovered model mechanism, revealing different intrinsic features controlling fine-grained aspects of generation, boosting further research on mechanistic interpretability of diffusion models.
The project website is at \href{https://foundation-model-research.github.io/difflens}{https://foundation-model-research.github.io/difflens}.
\end{abstract}    
\section{Introduction}
\label{sec:intro}

Diffusion models have emerged as powerful generative tools capable of generating high-quality content by iteratively denoising random noise  \cite{NEURIPS2020_4c5bcfec,song2021denoising}.
These models are now widely credited in applications such as content synthesis \cite{rombach2022high}, data manipulation \cite{kim2022diffusionclip}, and data augmentation \cite{melzi2023gandiffface}, impacting numerous aspects of human society.
However, diffusion models have been criticized for generating biased content \cite{luccioni2024stable,parihar2024balancing}, often resulting in skewed representations of sensitive social attributes such as gender, race, and age.
While biases may originate from the training data \cite{maluleke2022studying,perera2023analyzing}, they can be further amplified by the learned model itself \cite{perera2023analyzing}.
Given the extensive deployment of diffusion models, these biases may lead to detrimental effects in real-world contexts, such as reinforcing stereotypes in media representations or perpetuating inequalities in automated decision-making systems.

Existing approaches to debias diffusion models generally fall into two main strategies.
The first attempts to train unbiased models from scratch on biased datasets \cite{kim2024training} or fine-tune the target model \cite{shenfinetuning}, both of which are resource-intensive and can significantly impact the model's original performance.
The second aims to guide or edit the generation process \cite{parihar2024balancing,orgad2023editing,gandikota2024unified}, but these methods 
undermine the distinct internal workings of diffusion models,
risking overcorrecting the model behavior or distorting non-target attributes, which affecting generation quality.
In contrast, few studies have explored the intrinsic mechanisms within diffusion models that causally drive biased outcomes.

In this work, we hypothesize that neurons collectively play a pivotal role in driving the generation of biased concepts. 
These neurons exhibit specific features which we identify as intrinsic mechanisms within the model that contribute to social bias.
In particular, we set out to investigate these mechanisms to better understand and mitigate bias within diffusion models.

Our approach also aligns with the latest efforts on
understanding the inner workings of neural networks, which has 
drawn huge interest
within the community \cite{cheferhidden,park2024explaining}, not only due to the interpretability benefits that reveal model decision patterns \cite{konginterpretable,zhao2024explainability}, but also for its potential to enhance control over model behaviors \cite{bereska2024mechanistic,meng2022locating}.
However, the intrinsic neuron activities within diffusion models remain largely unexplored.
A significant challenge stems from the generative nature of diffusion models, which complicates, if not completely prohibits, fine-grained analysis at the neuron level.
The neuron polysemanticity \cite{bereska2024mechanistic}, which links individual neurons to multiple unrelated concepts additionally, further presents obstacles on mechanistic study of large-scale diffusion models.

To investigate the underlying mechanisms of bias generation in diffusion models, we present \method, a comprehensive framework that analyzes  the architectural components to identify the most influential elements within the model. 
Specifically, \method first disentangles the hidden neurons of the diffusion model into a sparse semantic feature space via a sparse autoencoder \cite{makhzani2013k}, which captures the patterns of neurons responsible for bias features such as gender and age.
Next, \method employs a gradient-based approach to pinpoint key components in the interpretable semantic space, identifying a minimal set of bias-specific features.
We can then further control or even mitigate the bias by adjusting the effects of these bias-related features.

\method offers two key advantages. 
First, via elaborately analyzing and identifying the inner mechanisms of diffusion process, it enables fine-grained control of bias 
with minimal impact on other semantic attributes.
Second, the identified bias features are intrinsic to the diffusion model, revealing its underlying contribution to biased generation.
Both have been demonstrated in our experiments.

In summary, our work makes several key contributions to the field of diffusion models.
(i) To the best of our knowledge, this is the first work to address bias mitigation in diffusion models through the lens of mechanistic interpretability, uncovering the underlying patterns of bias generation. Our method provides new insights into how specific neuron activities contribute to biased outcomes, offering a deeper understanding beyond standard latent space analysis.
(ii) We introduce a novel framework that includes semantic space construction, identification of influential features, and fine-grained control of bias generation based on the discovered features. 
This approach is not only effective for bias mitigation, but may also be applied to other aspects of diffusion model analysis.
(iii) We conduct extensive experiments on two pretrained diffusion models, demonstrating the effectiveness of our proposed mechanism discovery in comparison to several strong baselines for diffusion bias mitigation.

\section{Related Work}
\label{Sec:relate_work}

\noindent\textbf{Bias in Diffusion Models.}
Diffusion-based architectures often reflect and amplify social biases~\cite{luccioni2024stable,basu2023inspecting,rosenberg2023unbiased}, leading to potentially harmful outputs. 
Specifically, unconditional diffusion models exacerbate the biases present in the training data related to gender, race, and age~\cite{perera2023analyzing}. 
Text-to-image models, such as 
Stable Diffusion~\cite{rombach2022high},
tend to generate images of specific gender or stereotypes when prompted with occupational terms or terms with character traits and other descriptors~\cite{luccioni2024stable,cho2023dall,seshadri2024bias,bianchi2023easily}.

\noindent\textbf{Mitigating Diffusion Bias.}
Developing techniques to mitigate social biases in generative models has become increasingly crucial \cite{parihar2024balancing,li2024self}. 
Many current approaches focus on modifying prompts, such as adding gender-specific cues or ethical information, or using prompt learning techniques~\cite{friedrich2023fair,bansal2022well,kim2023de,chuang2023debiasing}. 
While these methods are effective for text-to-image systems, they are limited to scenarios involving textual prompts. 
Another strategy involves fine-tuning diffusion models using curated datasets~\cite{shenfinetuning}, which can effectively reduce bias but require extensive training and expensive high-quality data. 
Guided generation and model editing based 
methods~\cite{he2024debiasing,parihar2024balancing,orgad2023editing,gandikota2024unified} provide an alternative by introducing external guidance in the model's processing~\cite{he2024debiasing,parihar2024balancing} or manipulating attention weight~\cite{orgad2023editing,gandikota2024unified}. 
Although these techniques avoid retraining, they undermine the intrinsic mechanisms within the model specifically related to bias generation, often resulting in imprecise edits that inadvertently affect other parts of the image and degrade quality. 
In contrast, our approach allows for precise identification 
of bias-generating features, preserving overall image quality and maintaining other attributes, thereby providing a more reliable solution for bias mitigation.

\noindent\textbf{Interpretability of Diffusion Models.}
Most of the work interpreting diffusion models focus on explaining how the input affects the generation via attribution methods and attention maps ~\cite{hertzprompt,tang-etal-2023-daam,chefer2023attend,han2023svdiff}, 
but they do not reveal
the underlying mechanisms by which these attributes are processed and synthesized into a coherent output.  
Recent research has revealed that diffusion models possess a semantically rich latent space~\cite{kwon2023diffusion}, inspiring various studies on customized image editing through latent space manipulation~\cite{haas2024discovering, li2024self, park2023understanding}. 
Although these methods enable more flexible editing, they still do not provide a deep understanding of the model's inherent features, leading to unexpected results. 
Our approach directly interacts with the internal components of diffusion models, enabling us to identify and precisely modify key features like bias-specific ones. 
This allows for more accurate control and targeted editing within the model itself, also providing efficient tools for interpreting diffusion models. 
\section{Preliminary}
\label{Sec:preliminary}
\begin{figure*}[htbp]
\centering
\includegraphics[width=.95\textwidth]{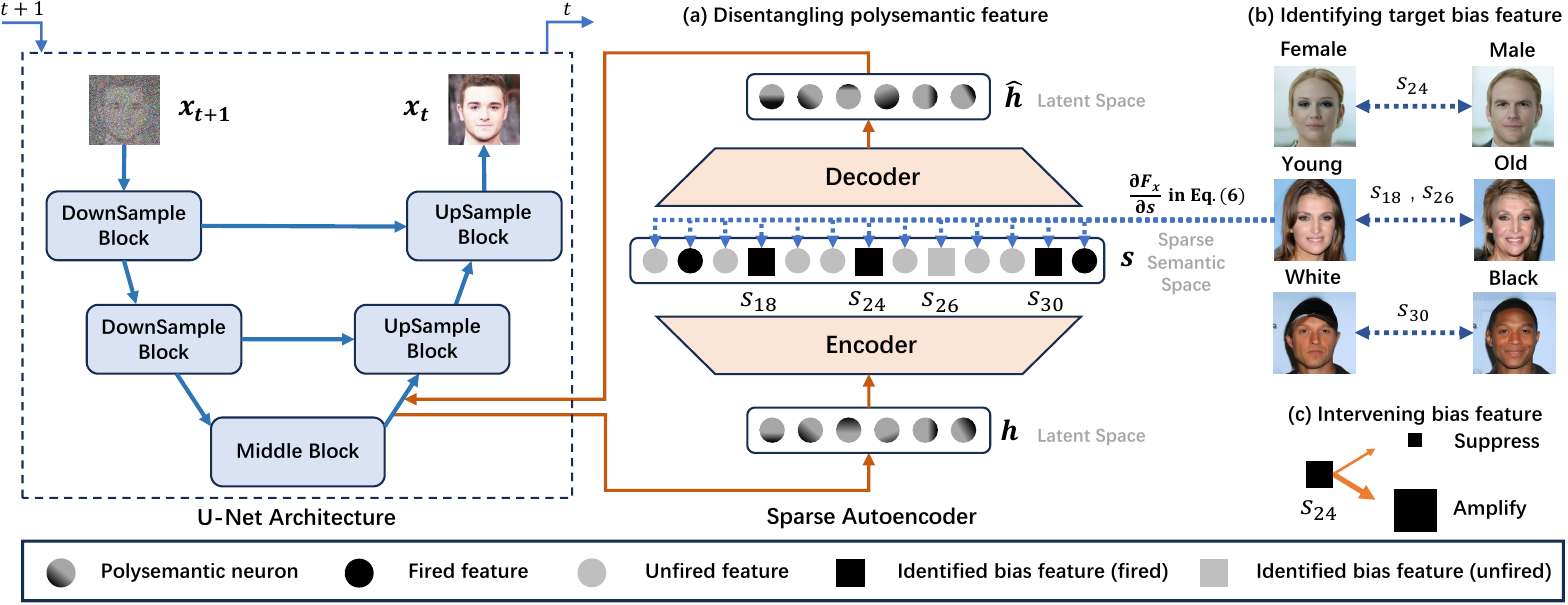}
\caption{
Framework of \method exploring the inner working of diffusion models for bias mitigation.
Neurons that are activated in the semantic space are defined as fired features and inactivated ones as unfired features.
Target features are either suppressed or amplified to control bias level.
We divide the framework into three parts (a), (b) and (c) which correspond to ~\cref{Sec:disentangle,Sec:identify,Sec:intervene}, respectively.
}
\label{figure:framework}
\vspace{-10pt}
\end{figure*}
In this section, we provide an overview of diffusion model architecture.
Diffusion models operate through two key processes, the forward and backward diffusion steps~\cite{NEURIPS2020_4c5bcfec}. 
Taking image generation as an example.
In the \textit{forward} process, noise is incrementally added to the input, transforming a real image \( \mathbf{x}_0 \) into pure noise over a series of timesteps \( t \in [0, T) \) where $T$ is the total number of timesteps. 
Mathematically, the forward process can be represented as 
\begin{equation}
    q(\mathbf{x}_t | \mathbf{x}_0) = \mathcal{N}(\mathbf{x}_t; \sqrt{\bar{\alpha}_t} \mathbf{x}_0, (1 - \bar{\alpha}_t) I)~.
\end{equation}
Here, \( \bar{\alpha}_t := \prod_{i=1}^t \alpha_i \) controls the amount of noise injected at each timestep \( t \), with \( \mathbf{x}_T \) becoming a near-random noise vector after the final timestep.
In the \textit{backward} process, the goal is to reverse this noising process and generate a clean image from the noisy sample. 
At each step, the diffusion network, \eg, U-Net \cite{ronneberger2015u}, models $ p_\btheta(\mathbf{x}_{t-1} | \mathbf{x}_t)$ with parameter $\theta$
predicting the denoised version given the current sample $ \mathbf{x}_t $. This reverse process is modeled as
\begin{equation}
    p_\btheta(\mathbf{x}_{t-1} | \mathbf{x}_t) = \mathcal{N}(\mathbf{x}_{t-1}; \mu_\btheta(\mathbf{x}_t, t), \Sigma_\btheta(\mathbf{x}_t, t))~,
\end{equation}
where \( \mu_\btheta \) and \( \Sigma_\btheta \) represent the mean and variance of the predicted denoised sample. 

Our work specifically focuses on the inner working of model itself, such as U-Net~\cite{ronneberger2015u} architecture in the diffusion model, to investigate the mechanisms underlying bias generation. 
We aim to examine the model's decision-making process through these representations to better understand the biases that emerge during image generation, providing deeper insights into the model’s internal functions and the precise factors driving its behavior~\cite{bereska2024mechanistic}.

\section{Methodology}
\label{Sec:method}
In this section, we propose \method, consisting of three stages aimed at uncovering internal mechanisms of bias generation in diffusion models and leveraging this understanding for bias mitigation.
Specifically, our goal is to discover the neuron activities that controls the generation of biased contents.
To achieve this, \method first disentangles 
the hidden states of neuron activation from the model layer into a sparse semantic space, capturing meaningful features that represent various attributes.
Next, it identifies the semantic features specifically related to biased content generation using a gradient-based attribution method.
Finally, by adjusting the scale of these identified features, \method allows for control over the level of bias in generated content, providing an effective approach for bias mitigation.
The overall framework has been illustrated in~\cref{figure:framework} and~\cref{alg:dissect,alg:bias_mitigation} in~\cref{Sec:appmethod}.

\subsection{Disentangling Polysemantic Neurons}
\label{Sec:disentangle}
We study the hidden states $\bh = [h_1, h_2, \ldots, h_n] \in \mathbb{R}^n$ from the diffusion model, representing the activations of the $n$ neurons at each network layer.
We assume that these neuron activations inherently capture the semantic information of the generation process.
Similar observations have been made in language models \cite{dai-etal-2022-knowledge,meng2022locating}, though little attention has been given to diffusion models in this regard.
\citet{kwon2023diffusion} demonstrated that diffusion models exhibit a semantic space within the U-Net~\cite{ronneberger2015u} architecture, specifically at the bottleneck layer. 
However, previous works have not precisely analyzed the different attributes within this space. 
In contrast, our method explicitly identifies semantic features and pinpoints the specific dimensions of neuron activations responsible for bias generation. 
This enables more precise and targeted control over bias, as demonstrated in our experiments in~\cref{figure:vissemantic,figure:contcurve}.

Note that, neurons are often \textit{polysemantic}, meaning that each neuron often corresponds to multiple, unrelated concepts, 
as discussed in \cite{olah2020zoom,mu2020,elhage2022toy}.
This places obstacles for understanding and interpreting the mechanism of the model, as the network's learned features are intertwined in a complex manner~\cite{bereska2024mechanistic}.
Nevertheless,
it also presents an opportunity that these features can potentially be disentangled by identifying directions in the activation space, where each activation vector can be expressed as a sparse linear combination of underlying features~\cite{huben2024sparse}. 
Thus, we leverage k-Sparse Autoencoders (k-SAE)~\cite{makhzani2013k} to analyze and interpret the semantic features encoded within the hidden states $\bh$.
It helps the method focus on salient features, disentangling polysemantic activations and isolating semantic attributes within the latent space, making it effective for understanding and manipulating key features in diffusion models.
Part (a) of ~\cref{figure:framework} has illustrated this processing stage.

Specifically,
let $\mathbf{H}^n = \{\bh\}$ be the original latent feature space with dimension number $n$.
We transform this feature space to a sparse high-dimensional space, denoted as \( \mathbf{S}^m = \{ \bs \}\) with dimension $m$ that is much larger than $n$.
In the space \( \mathbf{S}^m \), the features \( \bs \) 
are \textit{monosemantic}, \ie, each feature represents a distinct, non-overlapping semantic concept (discussions are  in~\cref{Sec:appmonosemanticityinsae}).
The mapping $\Phi : \mathbf{H}^n \rightarrow \mathbf{S}^m$ can be expressed by
\begin{equation}
    \label{Eq:encode}
    \bs = [s_1, s_2, \ldots, s_m] = \Phi(\bh) 
    = \text{TopK}(\bW_{\text{enc}}(\mathbf{h} - \mathbf{b}_{\text{pre}}))~.
\end{equation}
We map the vector $\bh$ to a sparse vector $\bs$ with $\|\mathbf{\bs}\|_0 = k$ and $k < m$, indicating that $\bs$ is sparse with only $k$ features activated.
This is achieved by the encoding process of a k-SAE
where 
$\bW_{\text{enc}} \in \mathbb{R}^{m \times n} $ and $  \bb_{\text{pre}} \in \mathbb{R}^{n} $ is the bias term.
$\text{TopK}(\cdot)$ is an operation that selects the top $K=k$ largest features, referred to as ``fired'' features, while setting the remaining features to zero, which are ``unfired'' features as shown in~\cref{figure:framework}.
For the decoding process, we have
\begin{equation}
    \hat{\mathbf{h}} = \Phi^{-1}(\bs) =\bW_{\text{dec}}\mathbf{s} + \mathbf{b}_{\text{pre}} ~,
    \label{Eq:decode}
\end{equation}
where $\hat{\mathbf{h}} \in\mathbb{R}^n$ is the reconstructed latent vector and $\bW_{\text{dec}} \in \mathbb{R}^{n \times m}$. 
$\Phi^{-1}$ represents the inverse mapping of $\Phi$ since $\bW_{\text{dec}}$ and $\bW_{\text{enc}}$ are initialized as transposes of each other.
We train the k-SAE using the loss function defined in \cref{eq:saeloss_function}, with the objective of minimizing the reconstruction error within the original hidden space, expressed as 
\begin{equation}\label{eq:saeloss_function}
    \mathcal{L}(\mathbf{h}) = \|\mathbf{h} - \hat{\mathbf{h}}\|_2^2 ~, 
\end{equation}
where $\mathbf{h}$ and $\hat{\mathbf{h}}$ refers to the original and the reconstructed hidden vectors, respectively.
We present results and implementation details 
in~\cref{Sec:apptrainsae}, demonstrating that we reconstruct the original latent space with few influences on the target model.
The encoding and decoding processes are presented in~\cref{alg:bias_mitigation} at~\cref{Sec:appmethod}. 

Disentangling the original latent space of diffusion model shares several advantages.
On one hand, transforming it into a sparse semantic space can well handle the polysemantic issue of the internal neurons within the neural network \cite{huben2024sparse}.
On the other hand, it enables more precise locating on the target feature for generation content attributes, such as identifying the bias features in the subsequent sections in our paper.
In addition, editing the whole state in the original hidden space like that in \cite{parihar2024balancing,kwon2023diffusion,haas2024discovering,li2024self} could inadvertently alter irrelevant attributes, caused by the polysemantic properties of the neuron in the model as shown by \cref{figure:viscomp,figure:vissemantic} in experiments.

\subsection{Identifying Bias Features}
\label{Sec:identify}
The next step is to identify the most influential features within the sparse semantic space that are associated with particular function, i.e., bias generation mechanism in this paper.
This need arises because accurately identifying the features that correspond to specific attributes can significantly enhance our ability to control and manipulate generated content responsibly.

Inspired by~\cite{sundararajan2017axiomatic,dhamdhere2018how}, we propose a gradient-based bias attribution method, to locate the features that correlates to the bias concepts.
We visualize this process in part (b) of~\cref{figure:framework}.
Intuitively, the neuron activations that influence the outcome of gender diverse in different aspects, which controls different characteristics such as hair style in~\cref{Fig:feature}.

Our goal is to measure the influence of the semantic feature $s_i \in \bs$ within $\mathbf{S}^m$ driving the biased contents 
on the target outcome.
We define $F_\bx : \mathbf{S}^m \to \mathbb{R}$ as a measure of interest, assessing the outcome of the model w.r.t. the target $\bs$ given the input $\bx$.
Specifically, we quantify the bias level as the probability of the social attribute class $y$ as $F_\bx(\bs)=\text{Pr}(y|\bs)$ regarding the disentangled feature vector $\bs$ derived from \cref{Eq:encode}.
Here $y$ represents different classes such as ``male'' or ``female'' in gender bias.
To locate the influential features, the internal mechanisms of diffusion model, we propose the attribution score of bias generation to the disentangled feature $s_i$ as
\begin{equation}
   S(s_i;\bx) = (s_i - s'_i) \cdot \int_{\alpha=0}^{1} \frac{\partial F_\bx(\bs' + \alpha (\bs - \bs'))}{\partial s_i} \, d\alpha ~,
    \label{Equa:sas}
\end{equation}
where $\bs'=[s'_1, \ldots, s'_m]$ is a relative baseline of $\bs$ and $\partial F_\bx(\bs' + \alpha (\bs - \bs'))/\partial s_i$ is the gradient of the bias measure given the input image $\bx$, to the target feature space $\bs$.
It is achieved by a light-weight classifier $F_\bx(\cdot; \bpsi)$ with parameter $\bpsi$.
The classifier construction follows \cite{parihar2024balancing} and
we provide more training details in~\cref{sec:apptrainhidden}.
Choices of baselines $s'$ are in~\cref{Sec:appalgodissect}. 

In \cref{Equa:sas}, the attribution score $S(s_i; \bx)$ aggregates the gradients along the path of the individual semantic feature $s_i$ ranging from zero effect ($\alpha=0$) to full effect ($\alpha=1$), given the input $\bx$, which is the generated content from the diffusion model.
If a feature $s_i$ significantly impacts the representation of a particular attribute, its corresponding gradient will be prominent, resulting in higher integrated values, \ie, higher attribution score $S$.

Further, to assess the overall influence of the semantic feature $s_i$ to the bias contents, we aggregate the attribution score as $S(s_i; X)=\sum_{j=1}^{N}S(s_i; \bx_j)$ where $\bx_j \in X$ with $X$ as the support dataset of $N$ generated samples.
\method then selects the top $\tau$ features with the largest attribution scores as \textit{bias features}, whose index set is
$\mathbf{A} = \{{i_1}, \ldots, {i_\tau}\}$, \ie, 
these are the most influential semantic features within the diffusion model driving the bias contents during generation.
The complete process of bias attribution is outlined in~\cref{alg:dissect} at~\cref{Sec:appmethod}. 
Notably, this step is required \textit{only once} for one specific diffusion model.

Additionally, we observe that the target can extend beyond $\bs$ to include the input $\bx$ or any component within the target model itself. This broader approach enables us to examine how these elements impact the behavior of the underlying model concerning the measure $F_\bx$.
To the best of our knowledge, we are the first to incorporate feature disentanglement and attribution methods to interpret the bias-related components within diffusion models, providing deeper insights into the internal workings of diffusion process.

\subsection{Intervening in Bias Features for Debiasing}
\label{Sec:intervene}
To control and mitigate social bias in diffusion models, we propose an effective method via intervening in the above discovered bias features indexed with $i \in \mathbf{A}$, where $\mathbf{A}$ consists of indexes of the identified bias features. 
It is possible to suppress or amplify these features to exert control over the model's behavior~\cite{dai-etal-2022-knowledge}. 
Unlike other methods~\cite{kwon2023diffusion,parihar2024balancing,li2024self}, we \textit{only} intervene in the identified features indexed with $\mathbf{A}$ rather than the entire semantic space.
Our intervention approach are illustrated in part (c) of~\cref{figure:framework}.

Specifically, we apply an intervention operation to the target features in the sparse semantic space.
Then these features are reconstructed into the original latent space for subsequent image generation and more details are in~\cref{Sec:appmethod}. 
For each feature indexed within $\mathbf{A}$ 
, we can either apply a scalar multiplication or adding an additive bias as 
\begin{equation} 
s_i = \text{Intervene}(s_i) =
\begin{cases}
     \beta s_i & (\text{Scaling}), ~ \text{or}\\
     s_i + \beta & (\text{Adding}). ~ \forall i \in \mathbf{A}, \beta \in \mathbb{R}. \quad \\
\end{cases}
\label{Eqa:alpha}
\end{equation}
By applying a distinct parameter $\beta$,
we can suppress or amplify the target bias feature to adjust the bias level or even mitigate it for improved balance.

Our \method demonstrates that it can control over the generation and representation of biased features in the output images as shown in~\cref{Sec:contedit}.
Furthermore, in \cref{sec:bias-mech}, we gain more valuable insights into how different bias features are embedded within the inner working of diffusion model, allowing us not only to debias effectively but also to better understand the model's decision-making processes. 
\begin{figure*}[htbp]
\centering
\includegraphics[width=1.0\textwidth]{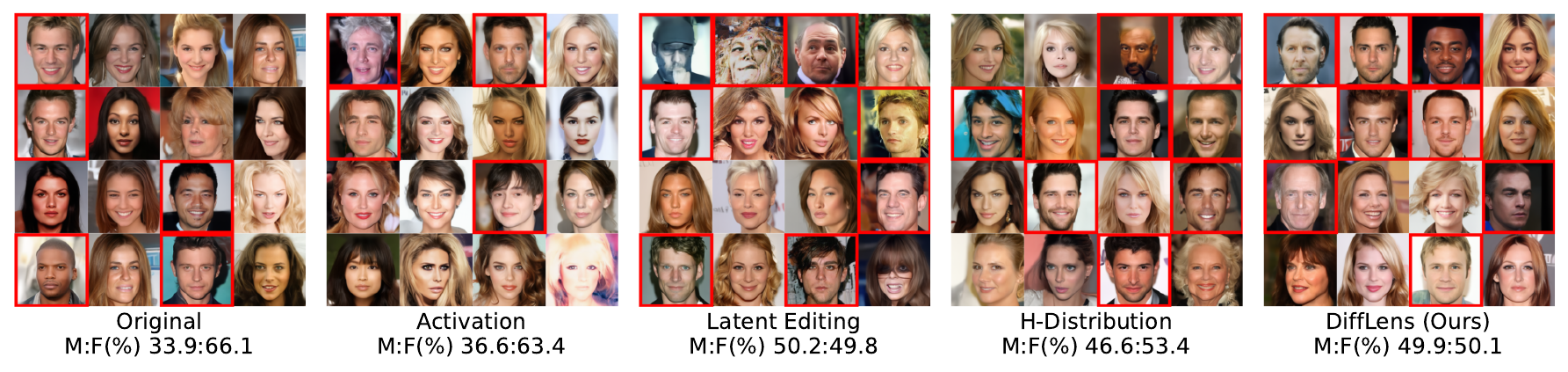}
\caption{
Comparison of randomly sampled original and debiased images from various baseline methods.
The minority group (male) is highlighted with red bounding boxes for easier viewing.
``M:F'' refers to the male-to-female ratio.
Our \method effectively mitigates bias while preserving high generation quality, whereas other methods either struggle to maintain balance or produce images with artifacts.
}
\label{figure:viscomp}
\end{figure*}

\begin{table*}[ht]
\centering
\resizebox{.9\textwidth}{!}{
    \begin{tabular}{@{}llccccccccc@{}}
    \hline
    \multirow{2}{*}{\textbf{Category}} &
    \multirow{2}{*}{\textbf{Method}} 
    & \multicolumn{3}{c}{\textbf{Gender~(2)}} 
    & \multicolumn{3}{c}{\textbf{Age~(3)}} 
    & \multicolumn{3}{c}{\textbf{Race~(4)}} \\ 
    \cmidrule(lr){3-5} \cmidrule(lr){6-8} \cmidrule(lr){9-11}
    &
    & \textbf{FD~$\downarrow$} & \textbf{FID~$\downarrow$} & \textbf{CLIP-I~$\uparrow$} 
    & \textbf{FD~$\downarrow$} & \textbf{FID~$\downarrow$} & \textbf{CLIP-I~$\uparrow$} 
    & \textbf{FD~$\downarrow$} & \textbf{FID~$\downarrow$} & \textbf{CLIP-I~$\uparrow$} \\ 
    \hline
    &Original & 0.226 & 33.38 & -
    & 0.592 & 33.38 & -
    & 0.718 & 33.38 & -\\
    \hline
    \multirow{2}{*}{\textbf{Guidance-based}} &Latent Editing~\cite{kwon2023diffusion}
    & \underline{0.003} & \textbf{29.85} & \underline{0.9474}
    & 0.606 & \underline{33.71} & 0.9092
    & \textbf{0.317} & \underline{34.61} & \underline{0.9081}\\
    & H-Distribution~\cite{parihar2024balancing} 
    & 0.048 & \underline{31.31} & 0.9440 
    & \underline{0.511} & 34.21 & 0.8594
    & 0.494 & 36.19 &  0.8762\\
    \hline
    \multirow{2}{*}{\textbf{Interpretability}} &Activation~\cite{bills2023language} 
    &0.190 & 34.27 & 0.8060
    & 0.544 & 48.91 & 0.7793
    & 0.700 & 46.13 & 0.7846 \\
     &\textbf{\method (Ours)} 
    & \textbf{0.002} & 31.93 & \textbf{0.9479} 
    & \textbf{0.401} & \textbf{31.71} & \textbf{0.9414}
    & \underline{0.447} & \textbf{33.47} & \textbf{0.9111}\\
    \hline
    \end{tabular}
}
\caption{Evaluation of bias mitigation in unconditional diffusion model P2~\cite{choi2022perception}.
The numbers beside each attribute indicate the number of classes for that attribute. 
We highlight the best results in \textbf{bold} and the second best with \underline{underlined} format (excluding the ``original'').
}
\label{Tab:uncond}
\vspace{-10pt}
\end{table*}
\section{Experiments}
In this section, we evaluate our \method qualitatively and quantitatively, following the below research questions.
\textbf{RQ1}: How effectively does \method mitigate social bias while maintaining image quality? 
\textbf{RQ2}: Can \method accurately identify intrinsic mechanism of bias generation?
\textbf{RQ3}: How well does it control the bias level through intervening in the found bias features?
\textbf{RQ4}: Is it able to interpret bias feature in a fine-grained way?

\subsection{Experimental Settings}
\label{Sec:experimental_setting}
\textbf{Diffusion Model Architecture.}
We evaluate our proposed method on two pretrained diffusion models, 
(i) P2~\cite{choi2022perception}, an \textit{unconditional} diffusion model trained on CelebA-HQ dataset~\cite{karras2018progressive};
(ii) Stable Diffusion~\cite{rombach2022high}, a text-\textit{conditional} diffusion model trained on LAION dataset~\cite{schuhmann2022laion}.

\noindent\textbf{Baselines.} 
We compare our method against several baselines that are categorized into interpretability and guidance-based methods.
For interpretability methods, we construct Activation~\cite{bills2023language} 
trying to locate influential neurons according to the neuron activation values. 
For guidance-based methods, 
we include Latent Editing~\cite{kwon2023diffusion} and H-Distribution~\cite{parihar2024balancing} (abbreviated as ``L.E.'' and ``H-Dist.''), trying to learn a latent vector to guide the unbiased generation within the bottleneck layer of U-Net. 
We additionally compare two methods, specifically for text-conditional diffusion models, which are Latent Direction~\cite{li2024self} and Finetuning~\cite{shenfinetuning}.
Due to Activation's limited debiasing effectiveness across both diffusion models and its restricted control over bias levels, we report its results only in~\cref{Tab:uncond} and \cref{figure:viscomp}.
More details about baselines are in \cref{Sec:appbaseline}.

\noindent\textbf{Implementation Details.}
To ensure fair comparison, and without loss of generality, 
we constrain our method, \method, and other benchmark methods to operate on the bottleneck layer of the U-Net~\cite{ronneberger2015u} in the diffusion model,
adhering to the baseline settings in \cite{kwon2023diffusion,parihar2024balancing}.
Notably, \method can easily extend the targeted mechanism to broader regions within the model. 
Following~\cref{Sec:method},
we train k-SAEs on the target latent space in the diffusion model.
For the training of the light-weight classifier, we follow the settings in~\cite{parihar2024balancing}.
For the support set of samples used for identifying bias feature, we leverage the diffusion models to generate 1,000 sample for each attribute. 
We provide more details of implementation and dataset we use 
in~\cref{Sec:appimple}.
The reproducible codes are in supplementary material.

\noindent\textbf{Evaluation Metrics.}
To measure the bias level of generated content,
we use the Fairness Discrepancy (FD) introduced in \cite{parihar2024balancing} which calculates the Euclidean distance between a reference distribution (typically uniform) and the bias distribution of generation, representing class distributions within an attribute (\eg, male and female for gender attribute). 
A lower FD score indicates more balanced outputs.
We also measure image quality using Fréchet Inception Distance (FID) \cite{heusel2017gans} on the generated images.
In addition, we follow~\cite{shenfinetuning} to measure 
(i) CLIP-T, which calculates the CLIP~\cite{radford2021learning} based semantic similarity between the generated image and the input text prompt, and 
(ii) CLIP-I, which assesses the similarity between originally generated images and images after debiasing.
More details are in~\cref{Sec:appexpsetting}.
\begin{table*}[ht]
\centering
\resizebox{\textwidth}{!}{
\begin{tabular}{@{}lcccccccccccc@{}}
\hline
\multirow{2}{*}{\textbf{Method}} 
& \multicolumn{4}{c}{\textbf{Gender~(2)}} 
& \multicolumn{4}{c}{\textbf{Age~(3)}} 
& \multicolumn{4}{c}{\textbf{Race~(4)}} \\ 
\cmidrule(lr){2-5} \cmidrule(lr){6-9} \cmidrule(lr){10-13}
& \textbf{FD~$\downarrow$} & \textbf{FID~$\downarrow$} & \textbf{CLIP-I~$\uparrow$} & \textbf{CLIP-T~$\uparrow$} 
& \textbf{FD~$\downarrow$} & \textbf{FID~$\downarrow$} & \textbf{CLIP-I~$\uparrow$} & \textbf{CLIP-T~$\uparrow$}
& \textbf{FD~$\downarrow$} & \textbf{FID~$\downarrow$} & \textbf{CLIP-I~$\uparrow$} & \textbf{CLIP-T~$\uparrow$} \\ 
\hline
Original & 0.564 & 120.06 & - &0.6155  &0.752 &120.06 & - &0.6155  &0.558 &120.06  & - & 0.6155 \\
Latent Editing~\cite{kwon2023diffusion} &0.408 & 166.11 & 0.8253 & 0.6005 &0.682  & 200.90 & 0.8527 & \textbf{0.6122} &0.524 & 153.05&\underline{0.8804} &0.6086\\
H-Distribution~\cite{parihar2024balancing} & 0.222
 & 151.68  &0.8475  & 0.6087 & 0.506  & 147.71
 & 0.8345
 & \underline{0.6098}
 & 0.544  & \underline{126.90}
&0.8255
 & 0.6100
\\
Latent Direction~\cite{li2024self}& 0.305 & \underline{129.37} & 0.8058& \underline{0.6091} &\underline{0.052}  & \underline{113.81} &0.8151  &0.6067 & \textbf{0.175} & 128.30 &0.8211 &\underline{0.6132}\\
Fintuning~\cite{shenfinetuning}& \underline{0.050} & 161.47 & \textbf{0.8779}& \textbf{0.6095}&0.746  & 161.47  & \textbf{0.8779}  &0.6095 & \underline{0.198} & 161.47&0.8779 &0.6095\\
\textbf{\method (Ours)} & \textbf{0.046} & \textbf{112.83}
 & \underline{0.8501} & 0.6090 & \textbf{0.049} & \textbf{99.17
} & \underline{0.8778} & 0.6057 & 0.401&\textbf{119.86
} & \textbf{0.9096}
 & \textbf{0.6149}\\
\hline
\end{tabular}
\vspace{-20pt}
}
\caption{
Evaluation of bias mitigation in text-to-image diffusion model Stable Diffusion~\cite{rombach2022high}, based on average performance across four occupations (see~\cref{Sec:appimple}).
We highlight the best results in \textbf{bold} and the second-best with \underline{underlined} text (excluding the ``original'').
}
\label{Tab:t2i}
\vspace{-5pt}
\end{table*}

\begin{figure*}[htbp]
\centering
\includegraphics[width=1.0\textwidth]{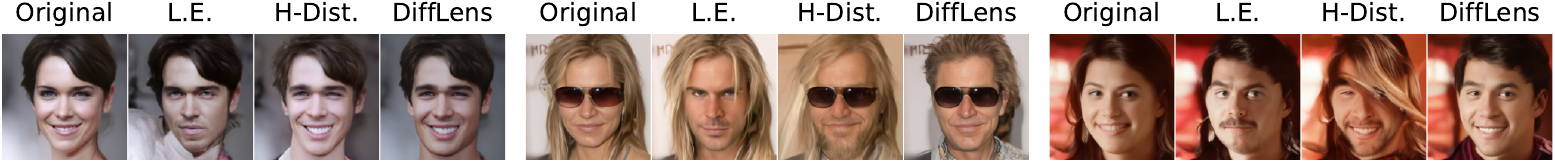}
\caption{
Comparison in accurate identification of bias features.
Our \method preserves overall image semantics such as smile and eyeglasses while other methods frequently introduce distortions or lose important details.
}
\label{figure:vissemantic}
\vspace{-10pt}
\end{figure*}

\subsection{Debiasing Diffusion Models}
\label{Sec:expdebias}
We define ``bias mitigation'' task as adjusting the diffusion model to produce balanced outputs across specific social attributes (\eg, gender). 
We focus on the task across gender, age, and race attributes. 
Gender includes male and female classes; age includes young, adult, and old; and race includes white, black, Asian, and Indian classes.

Intuitively, bias mitigation should yield balanced generation without compromising the overall semantics of the generated content.
To comprehensively evaluate the effectiveness of bias mitigation (\textbf{RQ1}) across different methods, we assess (i) the bias level using the FD score of generated images, (ii) the generation quality via the FID score, and (iii) the preservation of overall image semantics after bias mitigation 
using CLIP-I/T scores.
Beyond evaluating on the unconditional generative model, we also extend our analysis to a text-conditional one to showcase the broader applicability and generalizability of our method.

\noindent\textbf{Qualitative Results on Bias Mitigation.} 
We present batches of generated examples before and after debiasing gender attribute to visually evaluate bias level and generation quality,
as shown in~\cref{figure:viscomp}.
Our \method effectively neutralizes gender bias while ensuring the generated images remain realistic and intact.
In comparison, Latent Editing may generate unrealistic images, for example, the second one in the first row in~\cref{figure:viscomp}, H-Distribution struggles to generate a well-balanced output, and Activation produces imbalanced, distorted images. 
We also provide visual results for text-to-image diffusion model Stable Diffusion~\cite{rombach2022high} in~\cref{Sec:appvissd}.

\noindent\textbf{Quantitative Analysis on Bias Mitigation.}
We assess baseline methods on debiasing across various social attributes (\eg, gender) by generating balanced outputs. 
As shown in~\cref{Tab:uncond}, our \method demonstrates competitively strong results in (i) bias mitigation (measured by FD), (ii) in preserving semantic feature coherence (measured by CLIP-I) and (iii) maintaining generation quality (measured by FID).
Activation shows poor ability to mitigate social bias. 
Although Latent Editing performs well in mitigating gender bias and achieves favorable FID, it can produce images with artifacts, as shown in~\cref{figure:viscomp,figure:vissemantic}.
H-Distribution struggles to generate well-balanced outputs, as reflected by its poor FD scores. 

In~\cref{Tab:t2i}, we show results on a text-to-image  model, Stable Diffusion~\cite{rombach2022high}.
Our \method effectively mitigates bias, especially for gender and age, while effectively maintaining generation quality and image semantics 
among methods that are effective in bias mitigation.
Latent Direction achieves good debiasing for age and race but often overcorrects, reducing CLIP-I scores and compromising the balance between target generation ratio and image quality.
Finetuning, demanding extensive effort,  excels in gender and race but struggles with age bias mitigation.
Other baselines either handle multi-class attributes poorly or lack generalizability to text-to-image diffusion models.

\subsection{Accurate Identification of Bias Mechanism}
\label{Sec:accurate}
Accurate bias mechanism discovery could provide feasible control on the generated contents of diffusion, regarding the level of bias attributes.
For instance, unlike the original diffusion model P2~\cite{choi2022perception}, which generates a male-to-female gender ratio of approximately 3:7, our method easily attains the reverse ratio of 7:3. 
This capability goes beyond merely balancing generated images in \cref{Sec:expdebias}; it allows us to examine how effectively each method edits the target attribute, \eg, gender, while preserving non-target semantic features, \eg, eyeglass and smile (\textbf{RQ2}).
For illustration, we visualize the generated contents of compared methods upon the reversed ratio of gender attribute in \cref{figure:vissemantic}.
More examples and details are provided in~\cref{Sec:appaccurateattr}.
As is shown in~\cref{figure:vissemantic}, our method can preserve non-target attributes such as smile, background coherence, and even eyeglasses when debiasing, showcasing accurate identification of the bias generation mechanism by our method.
While the baseline methods often introduce distortions or compromise detail in these areas, without looking into inner working of bias generation.
Additionally, from ~\cref{Tab:uncond,Tab:t2i}, our \method consistently achieves high CLIP-I scores, demonstrating its ability to accurately identify and modify target bias specific features with minimal impact on unrelated content. 

\subsection{Fine-grained Bias Level Control and Editing}
\label{Sec:contedit}
\begin{figure}[htbp!]
\centering
\includegraphics[width=\linewidth]
{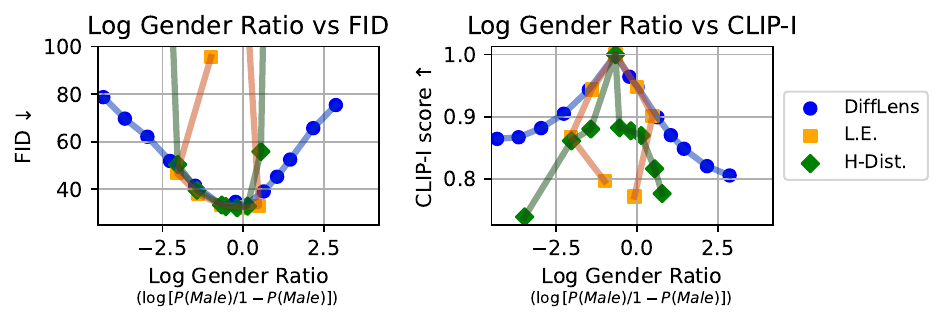}
\caption{
Achievable range of bias level control for each method.
The Log Gender Ratio reflects the log of male to female ratio in the generated images, with 0 indicating balance. 
Our \method offers broader bias control, delivering higher generation quality (left plot) and improved semantic coherence (right plot).
}
\label{figure:contcurve}
\vspace{-15pt}
\end{figure}

\begin{figure*}[htbp]
\centering
\includegraphics[width=1.0\textwidth]{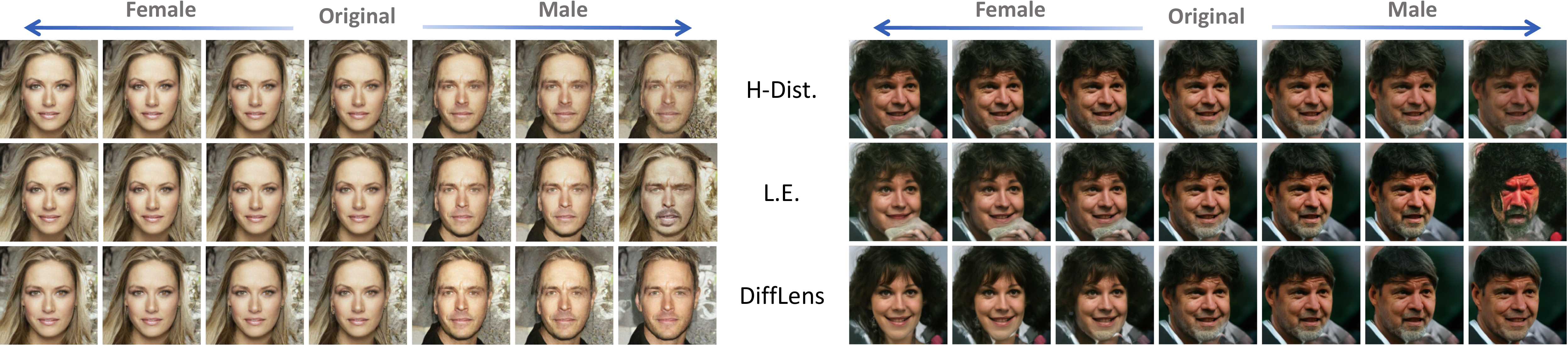}
\vspace{-10pt}
\caption{
Comparison of individual image transformations along the gender axis, with columns displaying images sampled from progressively shifting ratios as in~\cref{figure:contcurve}.
Our \method achieves smooth and consistent transitions that preserve semantic features like facial expressions, while other methods show distortions and loss of detail at higher imbalance ratios.
}
\label{figure:contvis}
\vspace{-10pt}
\end{figure*}

Benefiting from bias mechanism identification, we can control bias level with finer granularity and across a broader range (\textbf{RQ3}).
\Cref{figure:contvis} illustrates how different methods progressively modulate image generation in diffusion along two opposite gender directions.
More examples are in~\cref{Sec:appfinegrain}.
Our \method produces natural transitions,  consistently preserving semantic features such as facial expression and maintaining high generation quality. 
Examples about control in multi-attributes are in~\cref{Sec:appmulbias}.
The results of varying gender ratios generated by each method 
are shown in~\cref{figure:contcurve}.
Latent Editing and H-Distribution perform well when the gender ratio is around balanced, but their fidelity and visual coherence deteriorate at more skewed ratios, indicating a limitation in handling highly imbalanced outputs. 
By adjusting the value of intervention parameter incorporated in~\cref{Eqa:alpha}, our \method can flexibly control bias feature and generate different gender ratios, ranging from balanced to imbalanced ones (almost male or female), 
illustrating powerful controllability over the representations of bias.
Such highly imbalanced ratios also have wide applications in real world such as data augmentation for under-represented subgroups~\cite{parihar2024balancing}.

\subsection{Ablation Study}
\label{Sec:ablationStudy}
We conduct an ablation study to reveal how each component in our \method takes effect.
Recall that, it dissects the diffusion models by (i) disentangling the latent space to a sparse semantic space via k-SAE and (ii) identifying the bias features using bias attribution.
In~\cref{table:ablation}, directly selecting neurons with the highest activation values as bias features (Activation method) performs poorly, supporting the claim in \cref{Sec:disentangle}. Comparing ``Act@L'' with ``Act@S'' and ``Attr@L'' with \method, using k-SAE improves visual coherence, as seen in the higher CLIP-I scores (0.8597 to 0.9479 and 0.8060 to 0.9368), while avoiding direct edits on polysemantic neurons mitigates FID degradation (41.68 to 31.93). Furthermore, comparing ``Act@L'' with ``Attr@L'' and ``Act@'' with \method, bias attribution enables more precise localization of bias features, significantly lowering FD (0.190 to 0.072 and 0.303 to 0.002).
\method incorporating both shows superiority in bias mitigation with concise localization and intervention of bias features. 

\begin{table}[htbp!]
\centering
\resizebox{\columnwidth}{!}{
\begin{tabular}{lccccc}
\hline
\textbf{Components} & {\textbf{Orig.}} & \textbf{Act@L} & \textbf{Act@S} & \textbf{Attr@L} & \textbf{\method} \\
\hline
Activation & \ding{55} & \ding{51} & \ding{51} & \ding{55} & \ding{55} \\
K-SAE & \ding{55} & \ding{55} & \ding{51} & \ding{55} & \ding{51} \\Bias Attr. & \ding{55} & \ding{55} & \ding{55} & \ding{51} & \ding{51} \\
\hline \\[-0.7em]
\textbf{FD~$\downarrow$} & 0.226 & 0.190 & 0.303 & 0.072 & \textbf{0.002} 
\tikzmark{a}\\
\textbf{FID~$\downarrow$} &  33.38 & 34.27 & 32.70 & 41.68 & \textbf{31.93}  
\tikzmark{b}\\
\textbf{CLIP-I~$\uparrow$} & - & 0.8060 & 0.9386 & 0.8597 & \textbf{0.9479} 
\tikzmark{c}\\
\hline
\end{tabular}
}
\begin{itemize}
\small
    \item \tikz[baseline=(char.base)]{
        \node[circle, draw=cyan, fill=white, thick, minimum size=0.2cm, inner sep=0pt, text=cyan] (char) {\textbf{1}}; 
    },~\tikz[baseline=(char.base)]{
        \node[circle, draw=orange, fill=white, thick, minimum size=0.2cm, inner sep=0pt, text=orange] (char) {\textbf{2}}; 
    }~:~Bias attribution allows precise bias mitigation.
    \item \tikz[baseline=(char.base)]{
        \node[circle, draw=brown, fill=white, thick, minimum size=0.2cm, inner sep=0pt, text=brown] (char) {\textbf{3}}; 
    },~\tikz[baseline=(char.base)]{
         \node[circle, draw=purple, fill=white, thick, minimum size=0.2cm, inner sep=0pt, text=purple] (char) {\textbf{4}}; 
         }
    :~K-SAE improves visual feature coherence and maintains generation quality, via disentangling original latent space.
\end{itemize}
\vspace{-8pt}
\begin{tikzpicture}[overlay, remember picture]
    \draw[-stealth, thick, orange] (a) (3.3,2.0)--(3.3,1.9)--(0.6,1.9) -- (0.6,2.0) ;
    \node[circle, draw=orange, fill=white, thick, minimum size=0.1cm, inner sep=0pt,scale=0.7] at (2.55, 1.9) {\textcolor{orange}{\textbf{2}}};
    \draw[-stealth, thick, purple] (b) (3.3,1.3)--(3.3,1.2)--(1.85,1.2) -- (1.85,1.3) ;
    \node[circle, draw=purple, fill=white, thick, minimum size=0.05cm, inner sep=0pt, scale=0.7] at (2.55, 1.2) {\textcolor{purple}{\textbf{4}}};
   \draw[-stealth, thick, brown] (c) (3.3,1.65)--(3.3,1.55)--(1.85,1.55) -- (1.85,1.65);
    \node[circle, draw=brown, fill=white, thick, minimum size=0.1cm, inner sep=0pt,scale=0.7] at (2.55, 1.55) {\textcolor{brown}{\textbf{3}}};
    \draw[-stealth, thick, cyan] (c) (1.8,2.2)--(1.8,2.3)--(-0.6,2.3) -- (-0.6,2.2);
    \node[circle, draw=cyan, fill=white, thick, minimum size=0.1cm, inner sep=0pt,scale=0.7] at (0.6, 2.3) {\textcolor{cyan}{\textbf{1}}};
\end{tikzpicture}
\vspace{-8pt}
\caption{
Ablation study of \method in P2~\cite{choi2022perception} for gender attribute.
We abbreviate Original as ``Orig.'', neuron activations as ``Act.'', and bias attributions as ``Attr''. ``@L'' denotes operations on the original latent space, and ``@S'' on sparse semantic space.
}
\vspace{-15pt}
\label{table:ablation}
\end{table}
\subsection{Wider Investigation: Bias Mechanism}
\label{sec:bias-mech}
\begin{figure}[htbp!]
\centering
\includegraphics[width=0.85\linewidth]{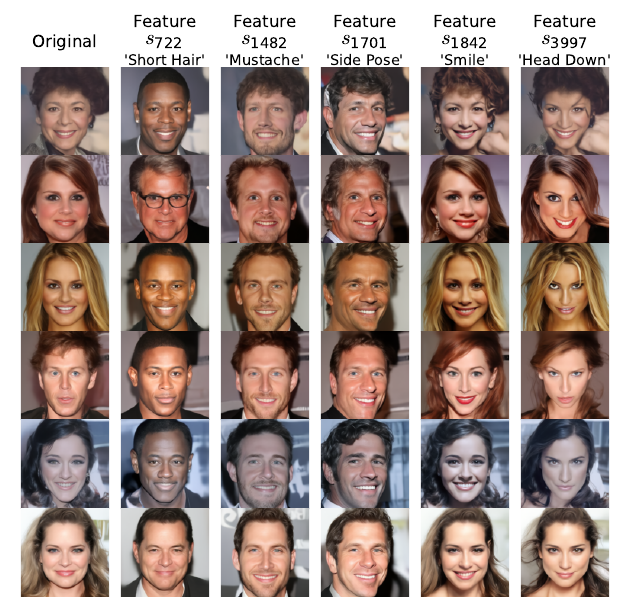}
\vspace{-10pt}
\caption{Manually analyzing the discovered bias features in diffusion models. Each column represents one specific feature $s_i, i\in \mathbf{A}$ and each row originates from the same raw image.
}
\label{Fig:feature}
\vspace{-15pt}
\end{figure}

We further look into the bias mechanism, \ie, the identified bias features $\{s_i|i \in \mathbf{A}\}$, to reveal its granular aspects on social attribute generation and how these features distinguishably affect the biased contents, shedding some lights on the inner workings of diffusion models (\textbf{RQ4}).
We observe from \cref{Fig:feature} that 
(i) multiple features may jointly influence a single social bias attribute, such as ``short hair'' and ``mustache'' contributing to gender perception; 
(ii) some features overlap across biases, for instance, ``side pose'' being associated with both age and gender biases as it often appears in images of older males; and 
(iii) poses like ``side pose'' and ``head down'' reflect in photography, often aligning with gender norms that men are depicted in dynamic or indirect poses, while women appear younger and more polished, reflecting social expectations~\cite{zhou2024bias,fairandbiasinaisci2024,zajko2021conservative}.
These stereotypes mirror underlying social biases, impacting not only how individuals are portrayed but also how their images are interpreted.
Our \method serves as a lens that uncovers social biases and stereotypes embedded in generated images, providing insight into how attributes such as gender, race, and age are systematically portrayed.

\section{Conclusion}
\label{Sec:conclusion}
In this work, we introduced a novel approach \method, for mitigating social biases in diffusion models by dissecting, analyzing and intervening in the internal mechanisms of the diffusion model. 
Leveraging k-SAE and gradient-based attribution method, we disentangle, identify, and control bias-related features with precision, allowing targeted bias mitigation while preserving non-target attributes. 
A wide range of evaluations across social attributes such as gender, age, and race validate effectiveness of \method in bias control and content preservation. 
Our work offers an interpretable solution to bias mitigation in diffusion models, though challenges remain, including the precise identification of certain racial features and a quality decline in intensified edits. 
Future directions include expanding this approach to other bias types, incorporating additional fairness perspectives, and exploring application across diverse model architectures like Diffusion Transformers~\cite{Peebles_2023_ICCV} for broader impact in ethical generative AI deployment.

\section{Acknowledgment}
The research received partial support from National Natural Science Foundation of China (Grant No. 62406193).
The authors also gratefully acknowledge further assistance provided by Shanghai Frontiers Science Center of Human-centered Artificial Intelligence, MoE Key Lab of Intelligent Perception and Human-Machine Collaboration, and HPC Platform of ShanghaiTech University.
{
    \small
    \bibliographystyle{ieeenat_fullname}
    \bibliography{main}
}
\clearpage
\setcounter{page}{1}
\maketitlesupplementary

\appendix
\section{Algorithm of \method}
\label{Sec:appmethod}
In this section, we present a detailed explanation of the methodology behind our \method. 
To enhance clarity, we outline the approach through two algorithms that comprehensively illustrate the key steps of \method.
In the context of the algorithms, ``online'' refers to performing the process during each image generation, while ``offline'' means executing it only once beforehand and using the obtained result directly during generation.
\subsection{Dissecting Bias Mechanism}
\label{Sec:appalgodissect}
We introduce how to dissect bias mechanism, \ie, disentangling bias features in the sparse semantic space (see in~\cref{Sec:disentangle} for more details).
\SetKwBlock{Begin}{}{end} 
\vspace{-0pt}
\begin{algorithm}
\caption{\method: Dissecting Bias Mechanism}
\label{alg:dissect}
\KwIn{A support set of samples $X = \{\bx_1, \ldots, \bx_N\}$, $\tau \in \mathbb{N}$}

\For{$j = 1$ \KwTo $N$}
{
    Extract $\mathbf{h} = [h_1, h_2, \ldots, h_n]$ from $\epsilon_\theta(\bx_j)$ \\
    $\bs = [s_1, s_2, \ldots, s_m] = \Phi(\mathbf{h})$ \tcp*[r]{\cref{Eq:encode}}
    \For{$i = 1$ \KwTo $m$}
    {
        $S(s_i;X) ~\texttt{+=}~ S(s_i;\bx_j)$ \tcp*[r]{\cref{Equa:sas}}
    }
}
$\mathbf{A} = \{i_1, \ldots, i_\tau\} = \operatorname{arg\,top}_{\tau} \{S(s_i;X) \mid s_i \in \mathbf{s} \}$\;
\KwOut{$\mathbf{A}$}
\end{algorithm}
\vspace{-0pt}

We denote U-Net~\cite{ronneberger2015u} as $\epsilon_\theta(\cdot)$ that accepts a sample $\bx$ as input and we use it to extract hidden representations of the sample.
To dissect bias mechanism, we leverage a support set of samples $X$ consisting of $N$ samples to identify bias feature.
For each sample, the original hidden state $\mathbf{h}$ of the sample is extracted and transformed into the sparse semantic space using a k-SAE~\cite{makhzani2013k} through~\cref{Eq:encode}.
Next, we localize the features that are related with bias contents by measuring the influence of these feature using a gradient-based bias attribution method as ~\cref{Equa:sas}.
For the calculation of attribution, we use Riemann approximation to estimate integral in the score of bias generation to the disentangled feature $s_i$ as follows
\begin{equation}
    S(s_i;\bx) = (s_i - s'_i) \cdot \cfrac{1}{q}\sum_{k=1}^{q} \cfrac{\partial F_\bx(\bs' + \cfrac{k}{q} (\bs - \bs'))}{\partial s_i} \,~,
    \label{Eq:attr_discrete}
\end{equation}
where $\bs'=[s'_1, \ldots, s'_m]$ is a relative baseline of $\bs$, $q$ is the number of discrete steps or partitions used in the Riemann approximation to estimate the integral, and $\partial F_\bx(\bs' + \alpha (\bs - \bs'))/\partial s_i$ is the gradient of the bias measure given the input image $\bx$, to the target feature space $\bs$.
The baseline $s'$ can be
(i) zero or (ii) a value tailored to a specific input.
We use (i) for unconditional diffusion model P2~\cite{choi2022perception} and (ii) for conditional diffusion model Stable Diffusion~\cite{rombach2022high}.
Then we aggregate the attribution scores across all sample and every time steps.
Finally, we set a threshold $\tau$ to the pick the $\text{top}~ \tau $ features that are highly related with specific bias content.

Note that, this process is required \textit{only} once for one specific diffusion model, since the sparse sematic space and the features are consistent across all time steps during generation and generalizable within a model.
As we utilize the same k-SAE~\cite{ronneberger2015u} across all time steps, thereby reducing the number of parameters, the diffusion step $t$ is not explicitly represented in~\cref{alg:dissect}.

\subsection{Bias Mitigation}
We describe how to intervene in bias features identified within the latent space of a diffusion model in~\cref{alg:bias_mitigation}. 
The process modifies specific elements of the semantic feature vector $\bs = [s_1, s_2, \ldots, s_m] \in \mathbb{R}^m$ according to 
a set of indices $\mathbf{A} = \{i_1, \ldots, i_\tau\}$ specifying the subset of features in $\bs$ corresponding to bias attributes to be adjusted.
\SetKwBlock{Begin}{}{end}
\vspace{-0pt}
\begin{algorithm}
\caption{\method: Bias Mitigating}
\label{alg:bias_mitigation}
\KwIn{
A set of feature indexes $\mathbf{A} = \{i_1, \ldots, i_\tau\}$, hidden state extracted from $\epsilon_\theta(\bx)$ $\mathbf{h} = \{h_1, h_2, \ldots, h_n\}\in\mathbb{R}^n$, $\beta \in \mathbb{R}$
}
$\bs = [s_1, s_2, \ldots, s_m] = \Phi(\mathbf{h})$ \tcp*[r]{\cref{Eq:encode}}
 Create $\bs' = \{s'_{1}, s'_{2}, \ldots, s'_{m}\} = \bs$ \\
\For{$k = 1$ \KwTo $\tau$}{
      $s'_{i_k} = \begin{cases}
         \beta s'_{i_k} & (\text{Scaling}), ~ \text{or}\\
         s'_{i_k} + \beta & (\text{Adding})\\
      \end{cases}$\tcp*[r]{\cref{Eqa:alpha}}
}
$\mathbf{h} = \mathbf{h} + \bW_{\text{dec}}(\bs' - \mathbf{s})$ \tcp*[r]{\cref{Eq:decode_trick}}
\KwOut{$\mathbf{h}$}
\end{algorithm}
\vspace{-0pt}

To map the intervened features back to the original hidden units, instead of directly reconstructing $\mathbf{h}$, we compute the difference between the intervened sparse semantic space $\bs'$ and the original space $\bs$, facilitating the mapping process.
This approach reduces reliance on the reconstruction effect of k-SAE~\cite{makhzani2013k}, 
as it only modifies the intervened parts while preserving the rest.
We reference for~\cite{huben2024sparse} and derive the operation as follows
\begin{equation}
\begin{aligned}
    \mathbf{h} &= \mathbf{h} + \hat{\mathbf{h}}' - \hat{\mathbf{h}} \\
    &= \mathbf{h} + \bW_{\text{dec}}\bs'+\mathbf{b}_{\text{pre}} - (\bW_{\text{dec}}\bs+\mathbf{b}_{\text{pre}})\\
    &= \mathbf{h} + \bW_{\text{dec}}(\bs' - \bs)\\
    &= \mathbf{h} + \sum_{k=1}^{\tau}(s'_{i_k}-s_{i_k})f_{i_k} ,
\end{aligned}
\label{Eq:decode_trick}
\end{equation}
where $\hat{\mathbf{h}}$ is the reconstructed hidden space without intervention in~\cref{Eq:decode}, $\hat{\mathbf{h}}'$ is the mapped back of intervened hidden state and $f_i$ is the $i_{\text{th}}$ row of the decoder matrix $\bW_{\text{dec}}$.
The resulting update is applied to the hidden state $\mathbf{h}$, effectively reflecting the feature adjustments within the model’s latent space.

By intervening on specific bias-related features, the algorithm enables controlled adjustments to the generated content.
We also provide two approaches for intervening, which are scaling and adding.
Since we adopt the same operation (scaling or adding) to hidden states $\mathbf{h}$ for all time steps, the time step $t$ is not explicitly represented in~\cref{alg:bias_mitigation}.
\section{Implementation Details}
\label{Sec:appimple}
In this section, we provide the details of how we implement \method.
To ensure consistency with the baselines used for comparison, we adopt the DDIM~\cite{song2021denoising} in both diffusion models (\cref{Sec:experimental_setting}) used in our experiment.
It modifies the original reverse process by allowing for non-Markovian updates, which reduces the number of timesteps needed for sampling without sacrificing image quality. 
The update rule for generating $ \mathbf{x}_{t-1} $ from $ \mathbf{x}_t $ can be written as
\begin{equation}
    \bx_{t-1} = \sqrt{\alpha_{t-1}} \, P_t(\epsilon_t^\theta(\bx_t)) + D_t(\epsilon_t^\theta(\bx_t)) + \sigma_t z_t,
\end{equation}
where $\alpha_t\in[0,1]$, $\sigma_t \geq 0, \sigma_t \in\mathbb{R}$ and $ z_t \sim \mathcal{N}(0, I)$. 
Here, $\epsilon_t^\theta$ is the predicted noise from the U-Net~\cite{ronneberger2015u}.
The intermediate terms are defined as:
\begin{equation}
    \label{Eq:pt}
    \begin{aligned}
    P_t(\epsilon_t^\theta(\bx_t)) &= \frac{\bx_t - \sqrt{1 - \alpha_t} \, \epsilon_t^\theta(\bx_t)}{\sqrt{\alpha_t}}, \\
    D_t(\epsilon_t^\theta(\bx_t)) &= \sqrt{1 - \alpha_{t-1} - \sigma_t^2 \, } \epsilon_t^\theta(\bx_t).
    \end{aligned}
\end{equation}
In this work, based on the finding of interference between $P$ and $D$ from~\cite{kwon2023diffusion}, we focus mainly on $P_t$, which corresponds to the prediction of the clean image at timestep $t$.

Additionally, we provide a strategy to intervene in the identified bias features for different classes within an attribute.
This approach supports both general bias mitigation and prompt-specific bias mitigation. 
For attributes with more than two classes (\eg, age or race), we intervene in the features of one class with a uniform probability $p = \frac{1}{k}$, where $k$ is the number of classes, while keeping the features of the remaining classes unchanged ($\beta$ = 1.0 for scaling or $\beta$ = 0.0 for adding in~\cref{Sec:intervene}).
For instance, in mitigating age bias, after identifying feature indices for ``young'', ``adult'' and ``old'', we intervene in the features of one of these three classes with a probability $p = \frac{1}{3}$, while leaving the features of the other classes unchanged within a batch of samples.
For bias mitigation in Stable Diffusion~\cite{rombach2022high}, the same strategy is applied to address varying levels of bias inherent to different prompts (\eg, ``doctor'' being male-dominated versus ``receptionist'' being female-dominated). 

We adopt the scaling operation to intervene in bias features for different attributes and specify the intervention parameter $\beta$.
For gender attribute, $\beta$ is set between 1.4 and 1.5. For the attribute, $\beta = 3.0$ is used for the old class, while $\beta = 1.0$ is applied to other classes. Similarly, for race attribute, $\beta = 5.0$ is applied to the black class, $\beta = 1.5$ for Asian and Indian, and $\beta = 1.0$ for white.
When mitigating bias, such as in the case of age, we use a uniform probability of $\frac{1}{3}$ to select one class per batch (\eg, old).
The selected class is intervened with its respective $\beta$ value (\eg, $\beta = 3.0$ for old), while keeping the other classes with $\beta = 1.0$. 
This approach ensures precise control and flexibility across various attributes and classes.

\subsection{Dataset Construction}
\label{Sec:appdataset}
We present details about constructing datasets using in (i) training k-SAE~\cite{makhzani2013k}, (ii) training the light-weight classifier for our \method and H-Distribution~\cite{parihar2024balancing}, and (iii) computing attribution score of bias generation.
For all three parts, we leverage the diffusion models (mentioned in~\cref{Sec:experimental_setting}) to generate image examples as our dataset.

\noindent\textbf{Dataset for Training K-SAEs.}
To construct the dataset for training k-SAEs, we leverage the unconditional diffusion model P2~\cite{choi2022perception} to generate 35,000 image examples.
For text-to-image diffusion model Stable Diffusion~\cite{rombach2022high}, we use prompt ``A face photo of a(an) {class} \{occupation\}'' to generate 100,000 images for training. 
The ``class'' represents the categories in the gender, age and race attributes, \eg, male.
The occupations are sampled from a prompt pool which is presented in~\cite{naik2023social}.

\noindent\textbf{Dataset for Training Light-weight Classifier.}
For the training of the light-weight classifier in the unconditional diffusion model P2~\cite{choi2022perception}, we use it to generate 1,000 images for each class in an attribute.
We use FairFace~\cite{karkkainen2021fairface} to determine the class for these generated images.
In text-to-image diffusion model Stable Diffusion~\cite{rombach2022high}, we use prompts ``a \{$class_1$\}\{$class_2$\}\{$class_3$\} person'', where the $class_i$ represent the categories in \{gender, age, race\} (\eg, ``a male old Indian person'').
We generate 1,000 images for each class in each attribute.

\noindent\textbf{Dataset for Identifying Bias Features.}
For the support set of samples $X$ used in identifying target bias features in~\cref{Sec:identify}, we generate 1,000 image samples for each attribute (gender, age and race) in both the unconditional and text-to-image diffusion model.
We use the trained light-weight classifier to discriminate the hidden space representations of these samples for  specific classes in an attribute.

\subsection{Training K-SAE on Hidden Space}
\label{Sec:apptrainsae}
In~\cref{Eq:encode,Eq:decode}, the k-SAE contains an encoder $\bW_{\text{enc}}$ and a decoder $\bW_{\text{dec}}$ with the same initialization of parameters.
For backpropagation through $\text{TopK}$ operation, we use straight through estimator. 
We use the DDIM~\cite{song2021denoising} to obtain the middle block representations in the U-Net~\cite{ronneberger2015u} for each image sample introduced in~\cref{Sec:identify}.
The loss we use is the reconstruction error introduced in~\cref{eq:saeloss_function}. 
The dimension $m$ of the sparse representation space is set to 4096 in P2~\cite{choi2022perception} and 5120 in Stable Diffusion\cite{rombach2022high}.
We train k-SAE in P2~\cite{choi2022perception} using 2 $\times$ RTX2080Ti GPU and in Stable Diffusion~\cite{rombach2022high} using 2 $\times$ A100 GPU.

\begin{figure*}[htbp]
\centering
\includegraphics[width=1.0\textwidth]{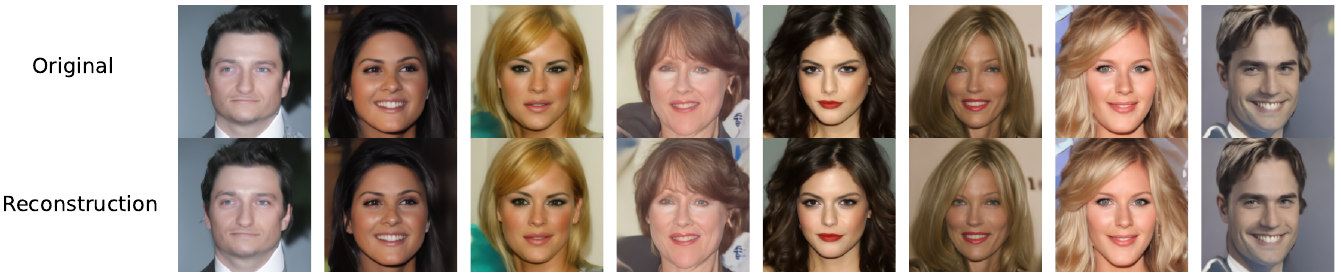}
\caption{
Comparison between images generated by original P2 model (top) and using k-SAE (bottom). The reconstructed and original images are almost identical, indicating effective reconstruction quality.
}
\label{figure:p2_reconstruction}
\end{figure*}

\begin{figure*}[htbp]
\centering
\includegraphics[width=1.0\textwidth]{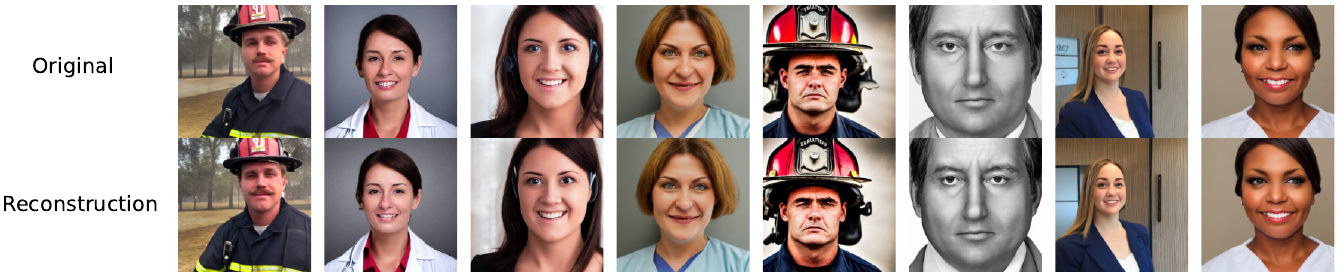}
\caption{
Comparison between images generated by original Stable Diffusion model (top) and using k-SAE (bottom). We observe minimal differences between the reconstructed and original images, indicating effective reconstruction quality.
}
\label{figure:sd_reconstruction}
\end{figure*}

\noindent\textbf{Reconstruction Effect with K-SAE.}
We present the effect of the reconstruction by using k-SAE in~\cref{Tab:reconstruction_quality}. 
The P2~\cite{choi2022perception} model is used to compare outputs with and without reconstruction and the non-reconstructed outputs is used as a reference dataset for both FID and CLIP-I metrics.
As we can see in~\cref{Tab:reconstruction_quality}, as $k$ goes larger, we obtain better reconstruction effect with the decreasing of FID and increasing CLIP-I.
However, the reconstruction is not the key factor, as we can see in~\cref{table:ablation_topk}, there is a trade-off between $k$ and our efficacy of bias mitigation.
We provide visual comparisons of reconstruction quality for P2~\cite{choi2022perception} and Stable Diffusion~\cite{rombach2022high} in~\cref{figure:p2_reconstruction,figure:sd_reconstruction}. Images reconstructed using k-SAE~\cite{makhzani2013k} are almost indistinguishable from those generated by the original diffusion model P2~\cite{choi2022perception}. 
For Stable Diffusion~\cite{rombach2022high}, using prompts detailed in~\cref{Sec:appevalpipe}, most semantic features are well reconstructed, although certain elements, such as the background, may not be fully preserved. 
This discrepancy is likely due to the richer semantic content of Stable Diffusion~\cite{rombach2022high} compared to P2~\cite{choi2022perception}.
Importantly, since our \method does not heavily depend on the reconstruction quality of k-SAE~\cite{makhzani2013k}, its performance in bias mitigation remains unaffected.
\begin{table}
\centering
\begin{tabular}{ccc}
\hline
$k$ & \textbf{FID~$\downarrow$} & \textbf{CLIP-I~$\uparrow$} \\
\hline
32 & 6.86 & 0.9516 \\
64 & 5.22 & 0.9695 \\
\textbf{128} & \textbf{3.71} & \textbf{0.9795} \\
\hline
\end{tabular}
\caption{Reconstruction effects w.r.t. different choices of $k$, where $k$ represents the number of activated features in k-SAE~\cite{makhzani2013k}. We use non-reconstructed (original) outputs as the reference dataset for calculating both FID and CLIP-I metrics.}
\label{Tab:reconstruction_quality}
\end{table}

\begin{figure}[H]
\centering

\includegraphics[width=\linewidth]
{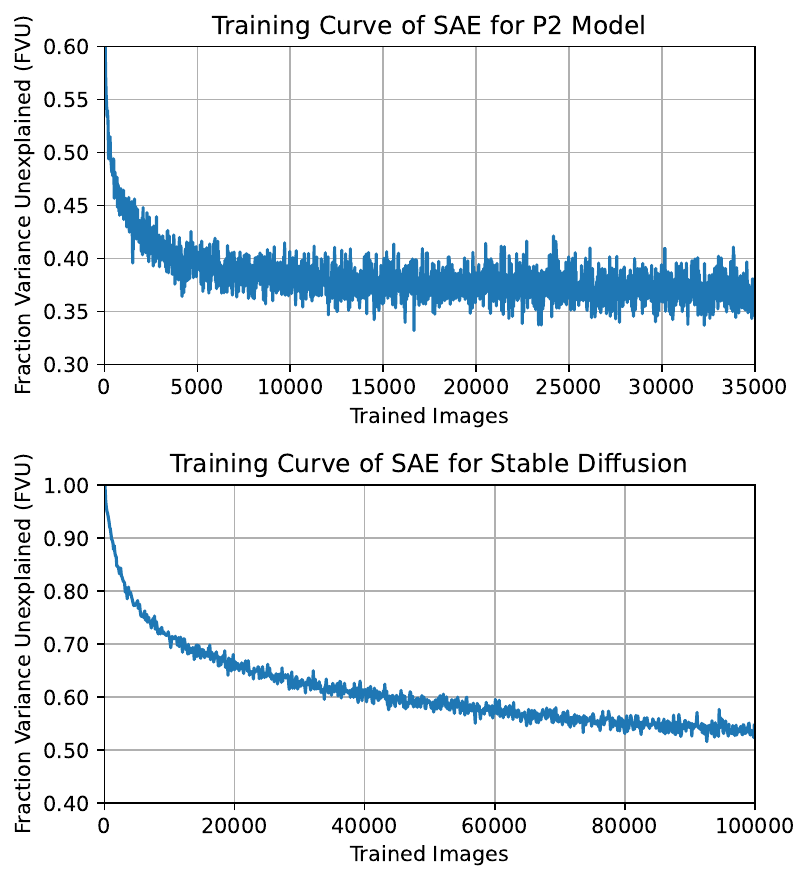}

\caption{Training curves of SAE for P2 Model and Stable Diffusion, showing the Fraction of Variance Unexplained (FVU) metric over the number of training images.}
\label{figure:training_curve}
\end{figure}

\noindent\textbf{Training Results.}
We present the details about the training of k-SAEs~\cite{ronneberger2015u}, including hyper-parameters and learning curves.
We train these on the hidden space without conditioning on the time step $t$, using the same model across all time steps. This approach significantly reduces the parameter size.
We use learning rates of 0.001 and 0.005, with batch sizes of 8 and 100, for P2~\cite{choi2022perception} and Stable Diffusion~\cite{rombach2022high}, respectively.
For the training curve, we leverage Fraction of Variance Unexplained (FVU)~\cite{makhzani2013k} which is a related metric of interest, measuring the total amount of the original activation that is not ``explained'' or reconstructed well by k-SAE~\cite{makhzani2013k}. 
We present it as below
\begin{equation}
FVU = \frac{\mathcal{L}(\mathbf{h})}{\text{var}[\mathbf{h}]}~,
\end{equation}
where $\mathbf{h}$ is the hidden state, $\mathcal{L}(\mathbf{h})$ is defined in~\cref{eq:saeloss_function} and var represents the variance of $\mathbf{h}$.
A lower FVU indicates better reconstruction performance, as more of the original activation is captured by the k-SAE model.
The training curves in~\cref{figure:training_curve} show that the k-SAE models for both P2~\cite{choi2022perception} and Stable Diffusion~\cite{rombach2022high} are trained to perform well on the FVU metric.

\subsection{Training Light-weight Classifier}
\label{sec:apptrainhidden}
\begin{figure}[htbp]
\centering
\includegraphics[width=\linewidth]
{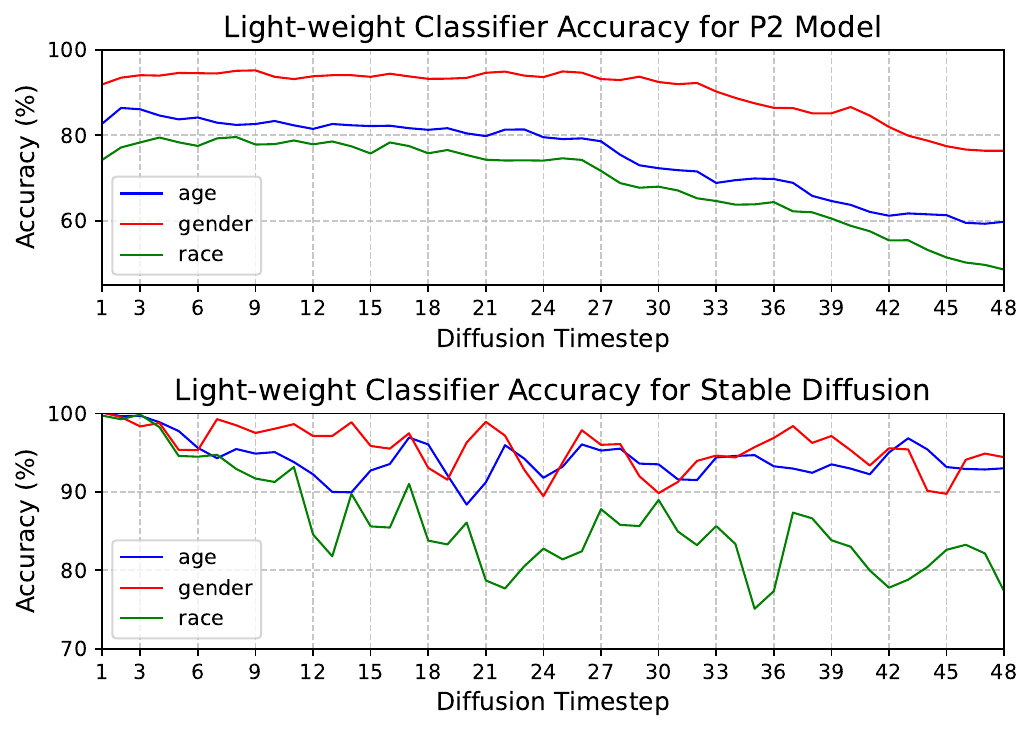}
\caption{Light-weight classifier accuracy for P2 Model and Stable Diffusion for all three attributes (gender, age and race). We follow~\cite{parihar2024balancing} and delete time time step 49. We use 1,000 images for each class in each attribute when training.}
\label{figure:h_classifier_accuracy}
\end{figure}

To accurately locate target features within the hidden space of diffusion models, we train a classifier to identify which feature are more closely related to categories within a given attribute (\eg, male and female for gender attribute) using the attribution method described in~\cref{Sec:identify}.
Here we give the details about how to train such a classifier.
After obtaining the dataset in~\cref{Sec:appdataset}, we follow the setting of~\cite{parihar2024balancing} for training.
Then we obtain the middle block hidden representations in the U-Net~\cite{ronneberger2015u} through DDIM~\cite{song2021denoising} with respective to all time steps.
The classifiers $ C_a^h(\mathbf{h}_t, t) $ are trained as linear heads over the obtained $ \mathbf{h}_t $, conditioned on time $ t $ and with respective to attribute $a$.
We train the classifier in both diffusion models using one RTX2080Ti GPU.
As shown in~\cref{figure:h_classifier_accuracy}, the classifiers for P2~\cite{choi2022perception} and Stable Diffusion~\cite{rombach2022high} achieve high accuracy across all three attributes: gender, age, and race. Specifically, the gender attribute comprises 2 classes, age consists of 3 classes, and race includes 4 classes. 
Notably, as the time steps approach the clean image state (closer to 1), the classifier accuracy consistently improves.

\section{Experimental Settings}
\label{Sec:appexpsetting}
\subsection{Evaluation Pipeline}
\label{Sec:appevalpipe}
In this section, we describe the overall steps for evaluation based on metrics in~\cref{Sec:experimental_setting}.
In unconditional diffusion model P2~\cite{choi2022perception}, we generate 10,000 images for each attribute, including gender, age, race, for each method.
We use FairFace~\cite{karkkainen2021fairface} as the classifier for different classes in all three attributes.
In text-conditional diffusion model Stable Diffusion~\cite{rombach2022high}, we use four prompts for evaluation. 
The prompt used is ``a face of an \{occupation\}'' where the ``occupation'' includes doctor, firefighter, nurse, and receptionist and two of which are male-biased and two female-biased. 
The first two is based on~\cite{parihar2024balancing} and the latter two prompts are from~\cite{naik2023social} which suggest bias for female and also exhibit age and race biases.
For every prompts, we generate 500 images for each method.
We will introduce how we measure FD, FID and CLIP-I/T in the subsequent sections.
\subsection{Evaluation Metric}
\label{Sec:appevalmetric}
\textbf{Fairness Discrepancy (FD).}
For evaluating bias mitigation, we use the metric FD in~\cite{parihar2024balancing}, and we provide details as follows. 
To measure the fairness of generated images with respect to a particular attribute $a$. 
Given a well-trained classifier for attribute $a$, denoted as $C_a$, the FD score is calculated as
\begin{equation}
    \|\bar{\mathbf{p}} - \mathbb{E}_{\mathbf{x} \sim p_{\btheta}(\mathbf{x})}[\mathbf{y}]\|_2~,
\end{equation}
where $ \bar{\mathbf{p}} $ is the target distribution over the attribute classes, which can be uniform, representing the ideal fair distribution, $ \mathbf{y} $ is the softmax output of the classifier $ C_a(\mathbf{x}) $ for the generated sample $ \mathbf{x} $, and $ p_{\btheta}(\mathbf{x}) $ is the distribution of the generated images.
For P2~\cite{choi2022perception}, we use the result of all images (10,000) in each method.
For Stable Diffusion~\cite{rombach2022high}, we average the results of four prompts, each with 500 images.

\noindent\textbf{Fréchet Inception Distance (FID).}
The Fréchet Inception Distance (FID)~\cite{heusel2017gans} is a metric commonly used to assess the quality and diversity of generated images in comparison to real images. 
Given two sets of images—one generated and one real—the FID measures the distance between the feature distributions of these two sets, as extracted by a pretrained Inception network. Let $ \mathcal{N}(\mu_r, \Sigma_r) $ and $ \mathcal{N}(\mu_g, \Sigma_g) $ represent the multivariate Gaussian distributions of real and generated images, respectively, where $ \mu_r $ and $ \Sigma_r $ are the mean and covariance of the real image features, and $ \mu_g $ and $ \Sigma_g $ are the mean and covariance of the generated image features.
The FID is defined as:
\begin{equation}
    \text{FID} = \|\mu_r - \mu_g\|^2 + \operatorname{Tr}\left(\Sigma_r + \Sigma_g - 2(\Sigma_r \Sigma_g)^{1/2}\right),
\end{equation}
where $ \|\mu_r - \mu_g\|^2 $ measures the squared difference between the means of the distributions, and $ \operatorname{Tr}\left(\Sigma_r + \Sigma_g - 2(\Sigma_r \Sigma_g)^{1/2}\right) $ measures the difference in covariances.

A lower FID score indicates a closer match between the distributions of real and generated images.
The reference dataset (real images) we use for unconditional diffusion model is CelebA-HQ dataset~\cite{karras2018progressive}, while for text-to-image diffusion models is the FFHQ~\cite{karras2019style}.
When calculating FID for Stable Diffusion~\cite{rombach2022high}, we evaluate images in all four prompts (2,000) simultaneously for each method.

\noindent\textbf{CLIP-I and CLIP-T.}
We follow~\cite{shenfinetuning} to measure the similarity between originally generated images and images after bias mitigation.
We use two metrics based on CLIP embeddings which are CLIP-I and CLIP-T scores. 
The CLIP-I score compares the similarity between the original image embedding $ \mathbf{e}_{\text{img}}^{\text{orig}} $ and the debiased image embedding $ \mathbf{e}_{\text{img}}^{\text{gen}} $, both extracted using the CLIP model. 
Formally, we compute the cosine similarity between $ \mathbf{e}_{\text{img}}^{\text{orig}} $ and $ \mathbf{e}_{\text{img}}^{\text{gen}} $ as
\begin{equation}
    \text{CLIP-I} = \frac{\mathbf{e}_{\text{img}}^{\text{orig}} \cdot \mathbf{e}_{\text{img}}^{\text{gen}}}{\|\mathbf{e}_{\text{img}}^{\text{orig}}\| \|\mathbf{e}_{\text{img}}^{\text{gen}}\|}.
\end{equation}
This score reflects the preservation of visual features after bias mitigation.

The CLIP-T score, on the other hand, measures the alignment between the debiased image and a text prompt. Given a text prompt embedding $ \mathbf{e}_{\text{text}} $ and the debiased image embedding $ \mathbf{e}_{\text{img}}^{\text{gen}} $, the CLIP-T score is calculated as
\begin{equation}
    \text{CLIP-T} = \frac{\mathbf{e}_{\text{text}} \cdot \mathbf{e}_{\text{img}}^{\text{gen}}}{\|\mathbf{e}_{\text{text}}\| \|\mathbf{e}_{\text{img}}^{\text{gen}}\|}.
\end{equation}
The CLIP model we use is CLIP ViT-L/14~\cite{radford2021learning}.
Note that, when we compare CLIP-I/T scores, we should also take FD score into account. 
The reason is that if one method is defective in bias mitigation, most of the time, it will not alter the generated images much, resulting in a high CLIP-I/T score.
Therefore, we mainly consider methods that are effective in bias mitigation when comparing CLIP-I/T score.
For CLIP-I in P2~\cite{choi2022perception}, we use the average result of every generated images (10,000) in each method.
For Stable Diffusion~\cite{rombach2022high}, we use the average result of the prompts in each method (\ie, calculating average CLIP-I/T for each image with a prompt and then average within four prompts).

\subsection{Baselines}
\label{Sec:appbaseline}
In this section, we provide details of the implementation of baseline methods along with specific information on each approach. 
For all baselines, we strictly follow their setting and record their results. 
All baselines except for Fintuning~\cite{shenfinetuning}, are based on the bottleneck layer of the U-Net~\cite{ronneberger2015u} in diffusion models.

\noindent\textbf{Activation} 
is based on previous work on interpreting GPT-2 using GPT-4 employing internal activation to analyze individual features~\cite{bills2023language}. 
It suggests that activation in internal neurons contains meaningful information.
In addition, \cite{dai-etal-2022-knowledge} shows the capability to edit internal neurons to affect the performance of a neural network.
Based on these, we adopt a similar setting where feature editing is performed directly on the original activation in the latent space. 
For a fair comparison, we constrain it on the bottleneck layer of the U-Net~\cite{ronneberger2015u}.

\noindent\textbf{Latent Editing}, as described in~\cite{kwon2023diffusion}, they learn a latent vector to steer unbiased image generation through the bottleneck layer of the U-Net~\cite{ronneberger2015u} model.
We adopt their approach to learn latent vectors for each class (\eg, ``old'') across different attributes (\eg, ``age''). 
To mitigate bias, we apply linear scaling to the learned vectors and incorporate them into the original bottleneck layer as described in their methodology.

\noindent\textbf{H-Distribution}, introduced by~\cite{parihar2024balancing}, employs distributional loss on bottleneck layer as guidance in diffusion models.
We directly use the h-classifier for gender attribute they provided, while we train the multi-class h-classifier for age and gender, following the methods introduced in the paper. 

\noindent\textbf{Latent Direction}, introduced by~\cite{li2024self}, identifies interpretable semantic directions within the latent space of text-to-image diffusion models, specifically in the U-Net~\cite{ronneberger2015u} bottleneck layer. 
This method optimizes concept-specific latent vectors by reconstructing images that exclude certain features in the text prompt while leveraging the pre-trained model's semantic knowledge.
We learn concept vectors for gender (male and female), age (young, adult, and old), and race (white, black, Asian, and Indian) following their setting.
These vectors are then combined with equal probabilities to mitigate bias.

\noindent\textbf{Finetuning}, introduced by~\cite{shenfinetuning}, 
uses distributional alignment loss  to adjust generated images to align with a user-defined target distribution. This approach integrates pre-trained classifiers to estimate class probabilities and fine-tune Stable Diffusion~\cite{rombach2022high}. 
The released fine-tuned model is utilized for generation.

\section{More Results}
\begin{figure*}[htbp!]
\centering
\includegraphics[width=1.0\textwidth]{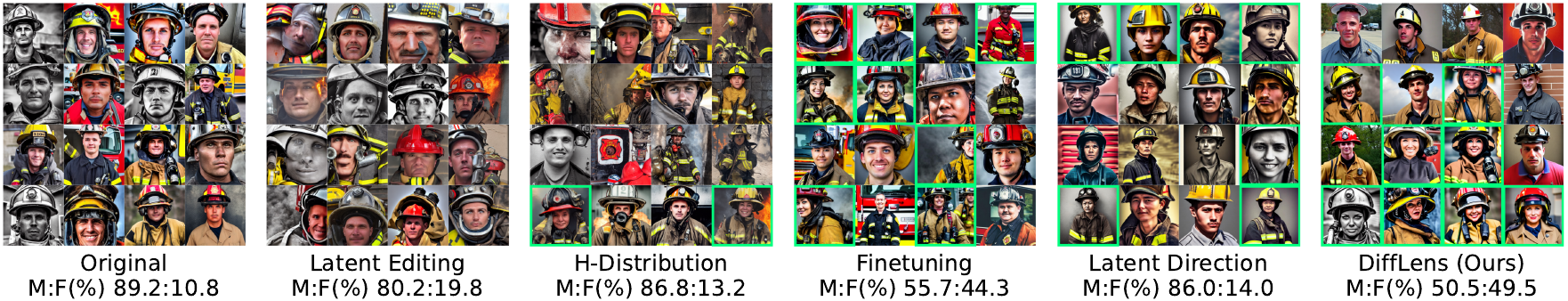}
\caption{
Comparison of randomly sampled original and debiased images generated by Stable Diffusion of the ``firefighter'' occupation.  Various baseline debiasing methods are compared. The minority group (female) is highlighted with green bounding boxes for easier viewing. "M:F" refers to the male-to-female ratio. 
}
\label{figure:viscompsd}
\end{figure*}
\subsection{Sensitivity Analysis of \textbf{\method}}
\label{Sec:appsensitivity}
In sensitivity analysis, we follow the experiment settings in~\cref{Sec:ablationStudy}, which aims at bias mitigation, \ie, generating balanced outputs.
We show that our \method is relatively stable within different hyper-parameters.

\noindent\textbf{K-SAE Hyper-parameters.}
We adjust different $k$ values to evaluate the effects of the k-SAEs~\cite{ronneberger2015u} on bias mitigation as shown in~\cref{table:ablation_topk}. As $k$ increases, it is harder to achieve a balanced outputs, though the reconstruction effect may be better. 
We conclude that out method does not heavily rely on how well we reconstruct the images (see in~\cref{alg:bias_mitigation} at~\cref{Sec:appmethod}), instead, how well the semantic features are disentangled and intervened plays a more important role. We choose $k$ with 32 showing the best performance in bias mitigation.

\noindent\textbf{Bias Attribution Parameters.}
We experiment on different number of target semantic features $\tau$ to be located. 
The results are summarized in~\cref{Tab:sensiig} where we selected $\tau = $ 30 features as the incorporated setting in our experiments, balancing both low FID of 31.93 and high CLIP-I score of 0.9479 in debiasing gender attribute. 
We observe that selecting too few features for bias mitigation risks omitting critical attributes related to biased content. Conversely, selecting a greater number of features facilitates better generation quality and semantic coherence, as evidenced by improvements in FID and CLIP-I metrics.
\begin{table}[htbp]
\centering
\resizebox{0.6\linewidth}{!}{
 \begin{tabular}{@{}cccc@{}}
 \hline
 \multirow{3}{*}{$k$} 
 & \multicolumn{3}{c}{\textbf{Gender~(2)}} \\ 
 \cmidrule(lr){2-4}
 & \textbf{FD~$\downarrow$} & \textbf{FID~$\downarrow$} & \textbf{CLIP-I~$\uparrow$} \\ 
 \hline
 Original & 0.226 & 33.38 & -\\
 \textbf{32} & \textbf{0.002} & \textbf{31.93} & 0.9479 \\
  64 & 0.005 & 31.94 & \textbf{0.9524} \\
 128 & 0.008 & 32.24 & 0.9113 \\
 \hline
 \end{tabular}
}
\caption{Impact of different TopK parameters in k-SAE on debiasing gender attribute. TopK means preserve the k features while deactivating the rest in the sparse semantic space (see in~\cref{Sec:disentangle}).}
\label{table:ablation_topk}
\end{table}

\begin{table}[htbp]
\centering
\resizebox{0.6\linewidth}{!}{
 \begin{tabular}{@{}cccc@{}}
 \hline
 \multirow{3}{*}{$\mathbf{\tau}$} 
 & \multicolumn{3}{c}{\textbf{Gender~(2)}} \\ 
 \cmidrule(lr){2-4}
 & \textbf{FD~$\downarrow$} & \textbf{FID~$\downarrow$} & \textbf{CLIP-I~$\uparrow$} \\ 
 \hline
 Original & 0.226 & 33.38 & -\\
 10 & 0.003 & 33.01 & 0.9446
 \\
 20 & \textbf{0.001} & 32.94
 & 0.9466 \\
 \textbf{30} & 0.002 & \textbf{31.93}
 & \textbf{0.9479} \\
 \hline
 \end{tabular}
}
\caption{Evaluation of bias mitigation for gender attribute \wrt various choices of feature number $\tau$, where $\tau$ means the number of identified target features in~\cref{Sec:identify}. We base on our choice of $k$ = 32 for this evaluation.}
\label{Tab:sensiig}
\end{table}
\subsection{Visual Results on Conditional Diffusion Model}
\label{Sec:appvissd}
We provide the visual results mentioned in~\cref{Sec:expdebias} for Stable Diffusion~\cite{rombach2022high}.
We include results for the prompt ``A face of a firefighter'' for illustration.
We use male-to-female ratio to measure how well the effect of bias mitigation is achieved for each method.
As shown in~\cref{figure:viscompsd}, our \method effectively mitigates bias while preserving generation quality. 
In contrast, Latent Editing~\cite{kwon2023diffusion} and H-Distribution~\cite{parihar2024balancing} struggle to produce balanced outputs and generate distorted images. 
While Finetuning~\cite{shenfinetuning} achieves high-quality images, it faces challenges in achieving well-balanced generation.

\subsection{Visual Results for Ablation Study}

\begin{figure*}[htbp!]
\centering
\includegraphics[width=1.0\textwidth]{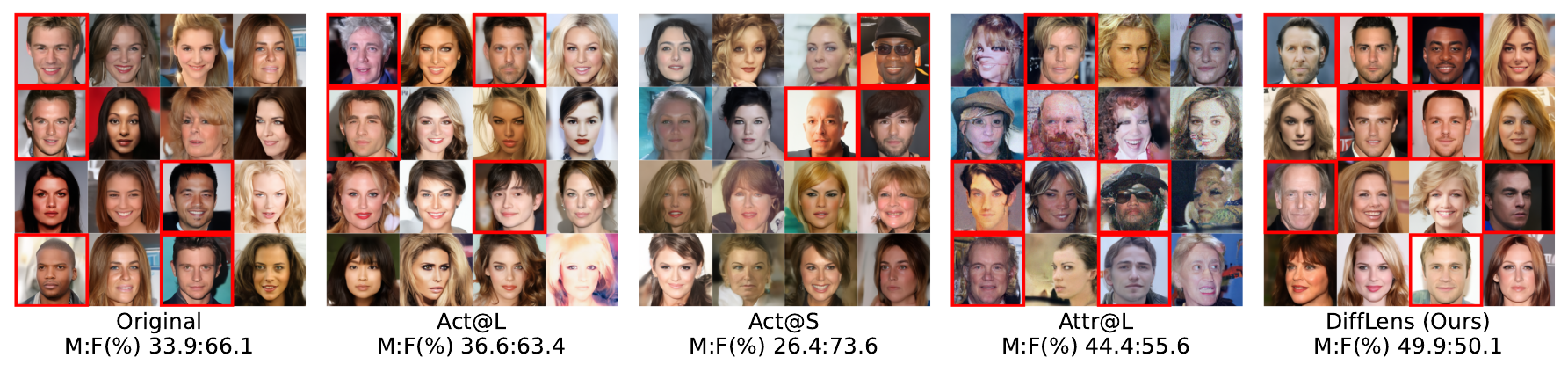}
\caption{
Qualitative results in ablation study of \method in P2~\cite{choi2022perception} for gender attribute.
We abbreviate Original as ``Orig.'', neuron activations as ``Act.'', and bias attributions as ``Attr''. ``@L'' denotes operations on the original latent space, and ``@S'' on sparse semantic space.
}
\label{fig:quliresablation}
\end{figure*}

We provide qualitative results for our ablation study in~\cref{Sec:ablationStudy}. 
As shown in~\cref{fig:quliresablation}, directly selecting neurons with the highest activation values as bias features (Activation method ``Act@L'') performs poorly, supporting the claim in \cref{Sec:disentangle}. 
Selecting features (disentangled by k-SAE~\cite{makhzani2013k}) according to activation value performs even worse (see in ``Act@S''). 
Comparing ``Attr@L'' with \method, although ``Attr@L'' has effect in debiasing, it suffers distorted outputs while \method are able to achieve excellent performance in debiasing and preserve image quality, illustrating the importance of disentangling neurons for better control. 

\subsection{Case Study}
\begin{figure*}[htbp]
\centering
\begin{subfigure}{1.0\textwidth}
    \centering
    \includegraphics[width=1.0\textwidth]{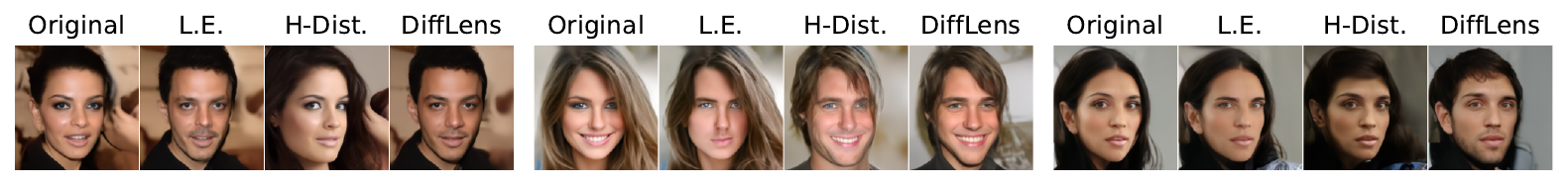}
\end{subfigure}

\vspace{5mm}

\begin{subfigure}{1.0\textwidth}
    \centering
    \includegraphics[width=1.0\textwidth]{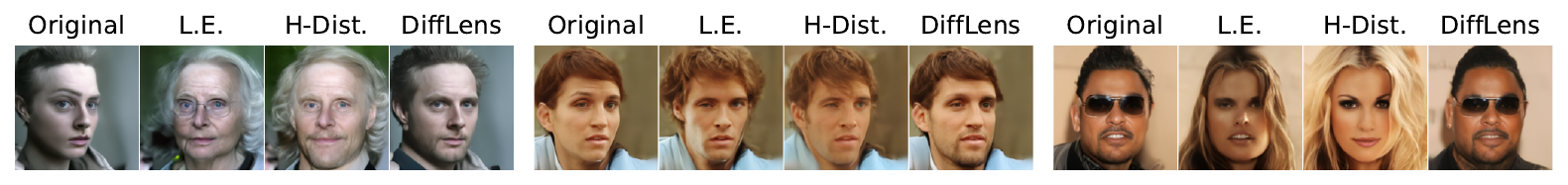}
\end{subfigure}

\vspace{5mm}

\begin{subfigure}{1.0\textwidth}
    \centering
    \includegraphics[width=1.0\textwidth]{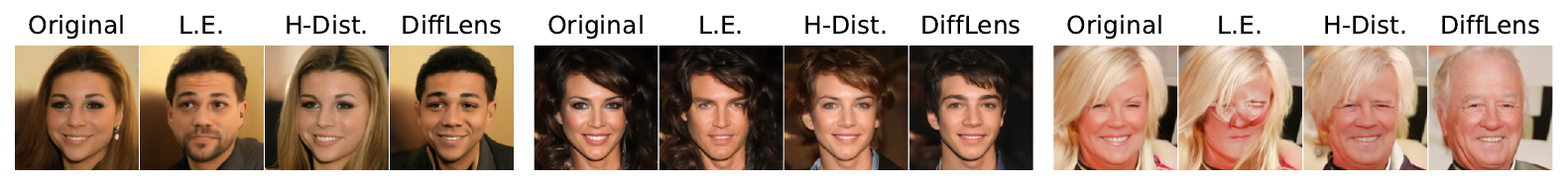}
\end{subfigure}

\vspace{5mm}

\begin{subfigure}{1.0\textwidth}
    \centering
    \includegraphics[width=1.0\textwidth]{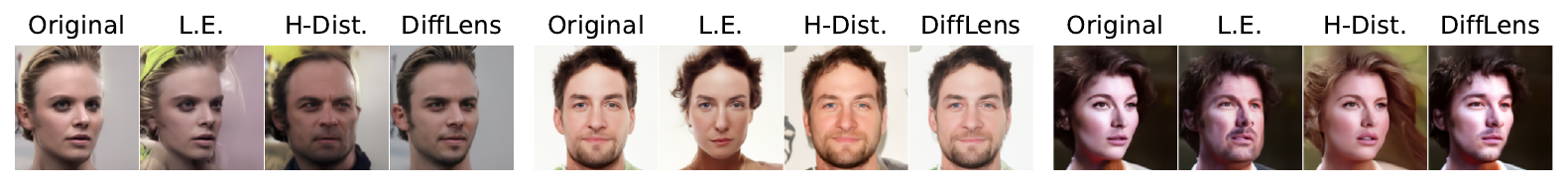}
\end{subfigure}

\vspace{5mm}

\begin{subfigure}{1.0\textwidth}
    \centering
    \includegraphics[width=1.0\textwidth]{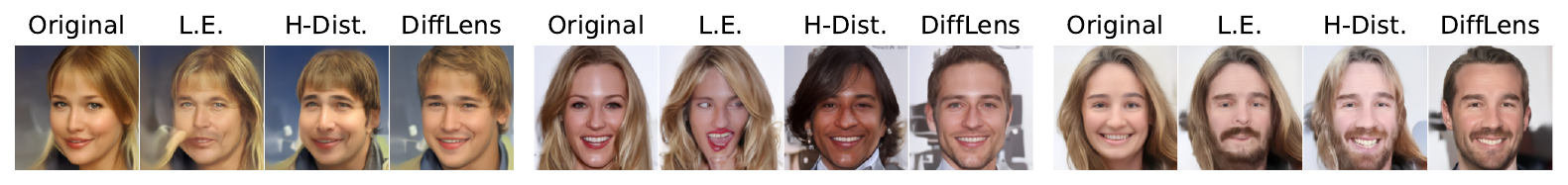}
\end{subfigure}

\vspace{5mm}

\caption{Comparison in accurate identification of bias features.
Our \method preserves overall image semantics such as smile and eyeglasses while other methods frequently introduce distortions or lose important details.}
\label{figure:app_d3_1}
\end{figure*}
\begin{figure*}[htbp]
\centering
\begin{subfigure}{.95\textwidth}
    \centering
    \includegraphics[width=1.0\textwidth]{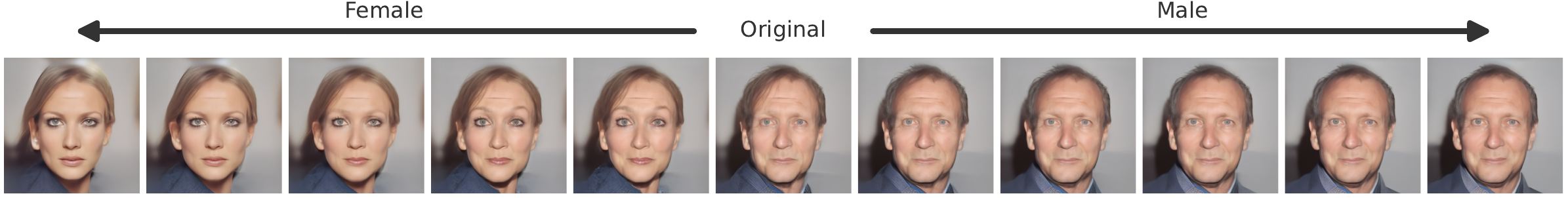}
\end{subfigure}

\vspace{5mm}

\begin{subfigure}{.95\textwidth}
    \centering
    \includegraphics[width=1.0\textwidth]{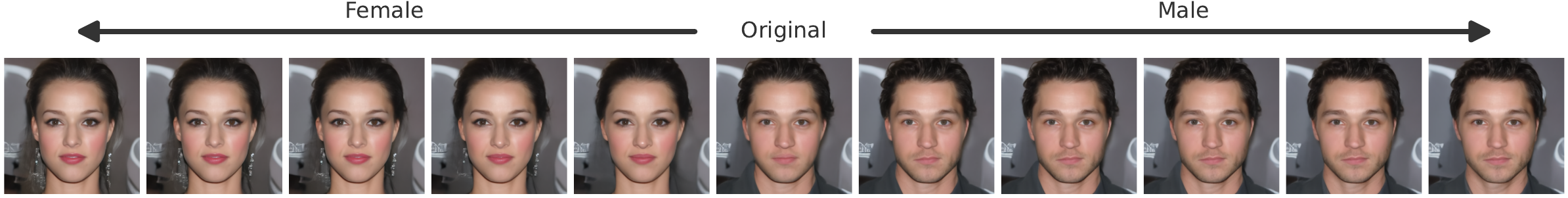}
\end{subfigure}

\vspace{5mm}

\begin{subfigure}{.95\textwidth}
    \centering
    \includegraphics[width=1.0\textwidth]{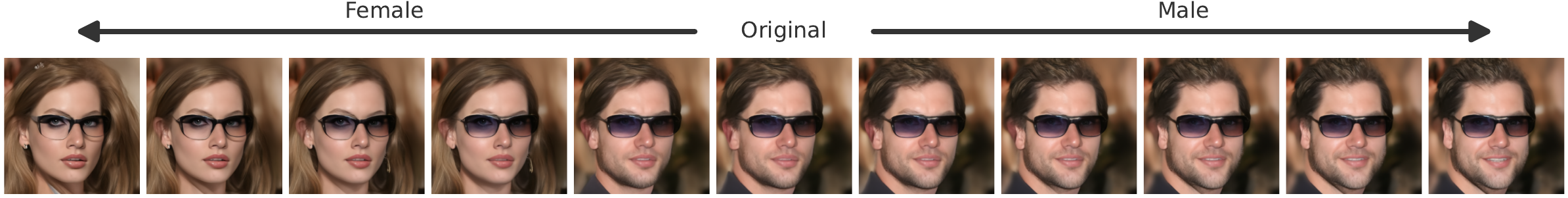}
\end{subfigure}

\caption{Scaling up ``female'' or ``male'' features. We adopt the same settings as outlined in~\cref{Sec:contedit}. Our \method demonstrates smooth editing and highlights its ability to exert fine-grained control over bias levels.}
\label{figure:Continuous_Gender_01}
\end{figure*}
\begin{figure*}[htbp]
\centering
\begin{subfigure}{.95\textwidth}
    \centering
    \includegraphics[width=1.0\textwidth]{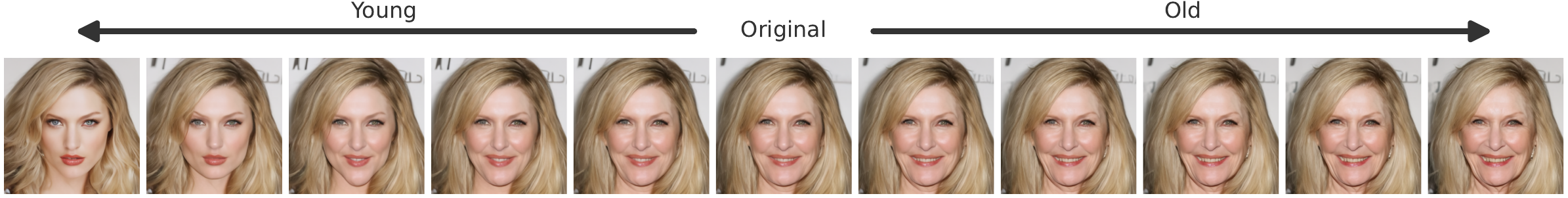}
\end{subfigure}

\vspace{5mm}

\begin{subfigure}{.95\textwidth}
    \centering
    \includegraphics[width=1.0\textwidth]{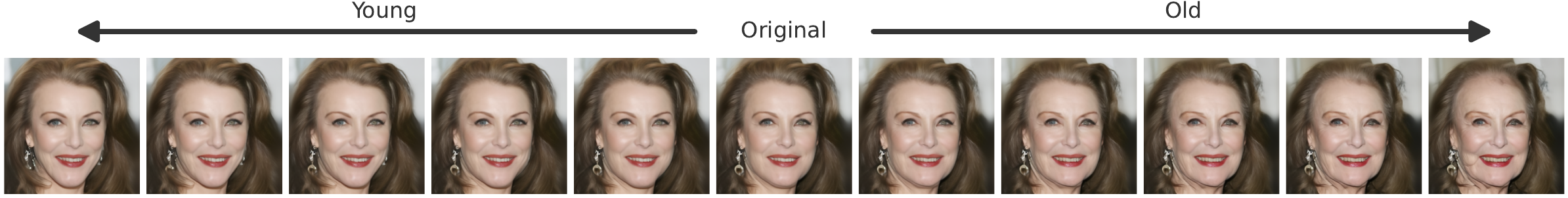}
\end{subfigure}

\vspace{5mm}

\begin{subfigure}{.95\textwidth}
    \centering
        \includegraphics[width=1.0\textwidth]{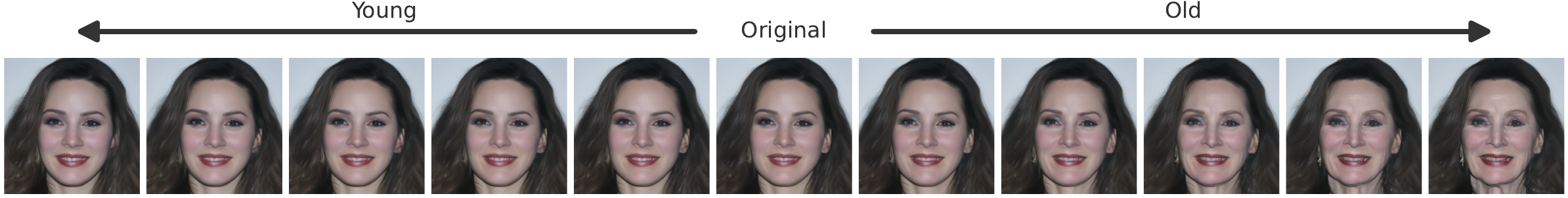}
\end{subfigure}

\caption{Scaling up ``young'' or ``old'' features. We adopt the same settings as outlined in~\cref{Sec:contedit}. Our \method demonstrates smooth editing and highlights its ability to exert fine-grained control over bias levels.}
\label{figure:Continuous_Editing_Age}
\end{figure*}
\begin{figure*}[htbp]
\centering
\begin{subfigure}{.95\textwidth}
    \centering
    \includegraphics[width=1.0\textwidth]{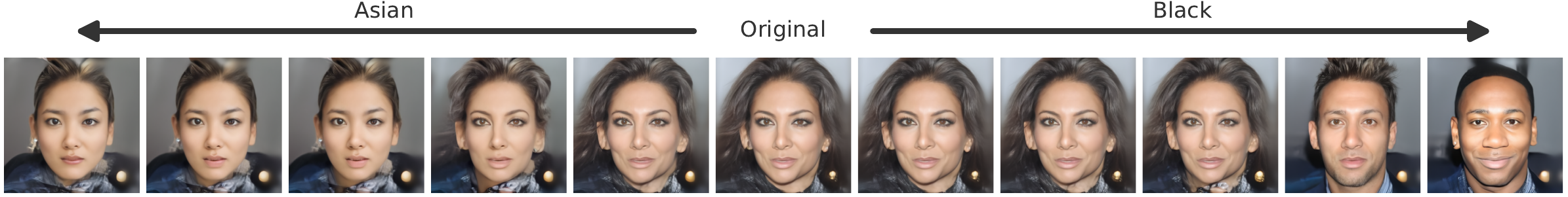}
\end{subfigure}

\vspace{5mm}

\begin{subfigure}{.95\textwidth}
    \centering
    \includegraphics[width=1.0\textwidth]{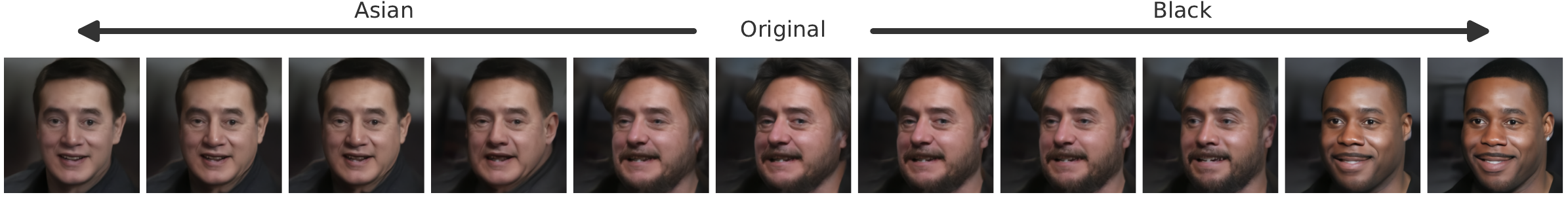}
\end{subfigure}

\vspace{5mm}

\begin{subfigure}{.95\textwidth}
    \centering
    \includegraphics[width=1.0\textwidth]{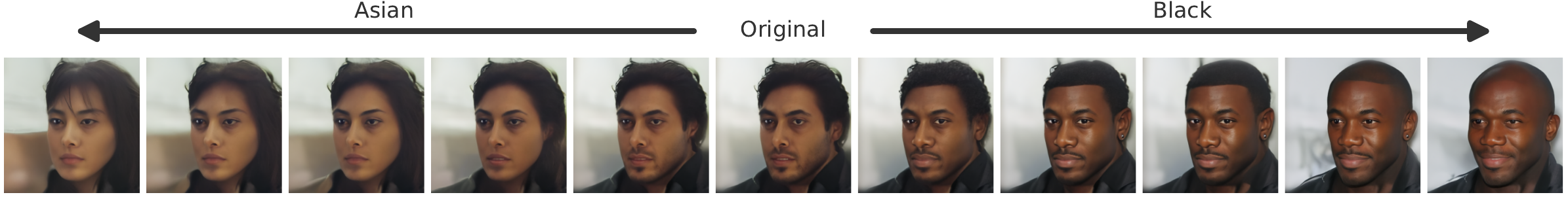}
\end{subfigure}

\caption{Scaling up ``Asian'' or ``Black'' features. We adopt the same settings as outlined in~\cref{Sec:contedit}. Our \method demonstrates smooth editing and highlights its ability to exert fine-grained control over bias levels.}
\label{figure:Continuous_Editing_Race}
\end{figure*}
\subsubsection{Accurate Attribution of Bias Features}
\label{Sec:appaccurateattr}
We provide more examples with our \method corresponding to ~\cref{Sec:accurate}.
\Cref{figure:app_d3_1} provides visual examples for comparison with baseline methods.

Following the same settings for generating a male-to-female ratio of 7:3 as described in~\cref{Sec:accurate}, our \method effectively preserves semantic features unrelated to the target attribute (gender), such as facial expressions, eyeglasses, background, and race.

In contrast, H-Distribution~\cite{parihar2024balancing} may unintentionally alter non-target attributes, such as race. For instance, in the left example of the first row and the middle example of the last row, the generated images exhibit changes in racial attributes. Additionally, the background in their generated images may vary significantly, as seen in the middle and right examples of the fourth row.
Latent Editing~\cite{kwon2023diffusion}, on the other hand, may generates distorted or unrealistic images, such as the left example from the last row. It also tends to entangle attributes like age and gender, as demonstrated in the left example of the second row, where a male image appears to incorporate old-age features.
For original images that are male, in the target 7:3 male-to-female ratio, our \method can maintain these images almost unchanged (e.g., the middle example in the fourth row and the right example in the second row in~\cref{figure:app_d3_1}), aligning with the desired ratio. In contrast, other methods may inadvertently alter the original images or transform them into female representations, failing to preserve the intended male attributes.
\begin{figure*}[htbp]
\centering
\begin{subfigure}{.95\textwidth}
    \centering
    \includegraphics[width=1.0\textwidth]{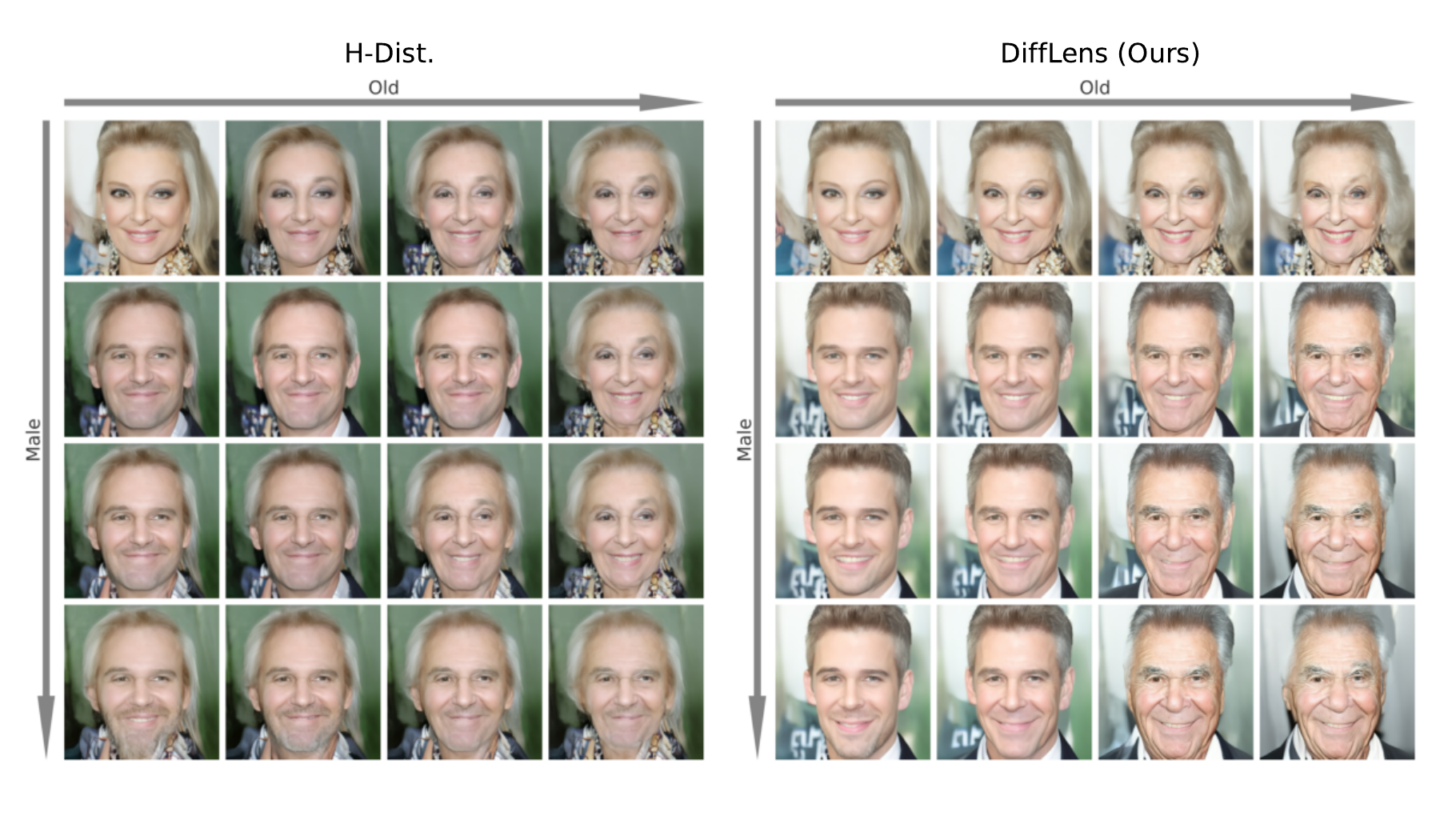}
\end{subfigure}

\begin{subfigure}{.95\textwidth}
    \centering
    \includegraphics[width=1.0\textwidth]{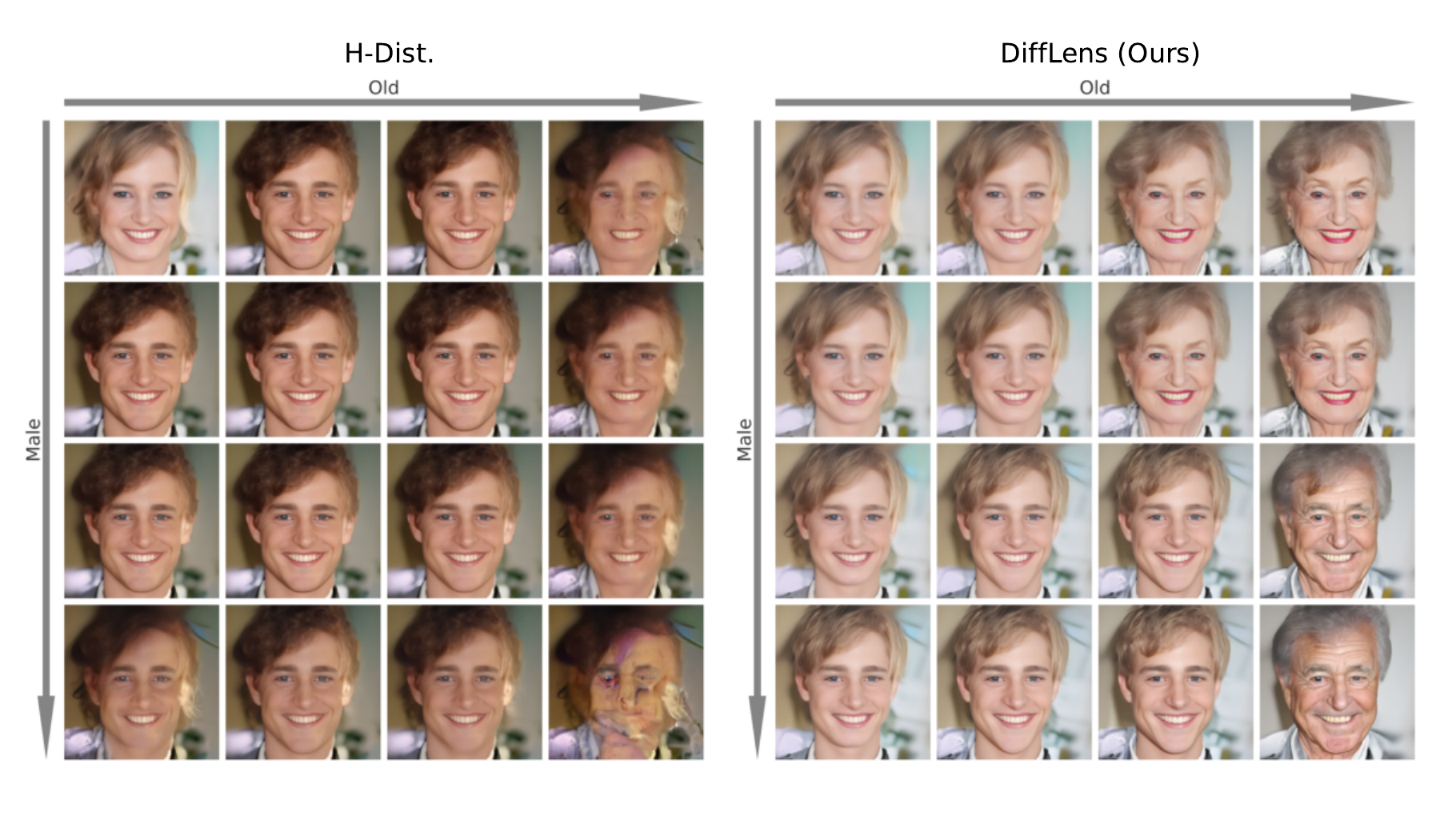}
\end{subfigure}

\vspace{-20pt}
\caption{Scaling up ``old'' and ``male'' features at the same time. We try to control over the bias level across multi-attribute simultaneously rather than a single attribute, illustrating our ability to disentangle different bias features and accurate identification of these features. We provide two examples for control over age and gender attributes.}
\label{figure:Gender_Age_01}
\end{figure*}
\subsubsection{Fine-grained Control and Editing}
\label{Sec:appfinegrain}
More examples showing our control over bias level with finer granularity (see in~\cref{Sec:contedit}) are provided within this section.
The settings outlined in~\cref{Sec:contedit} are applied, where generated images are randomly sampled across a broad spectrum of ratios (\eg, male-to-female and young-to-old).

We release the results of how we transform the original image along two gender directions in~\cref{figure:Continuous_Gender_01}.
As we can see in~\cref{figure:Continuous_Gender_01}, when editing towards male or female, we preserve generation quality and the visual feature coherence such as eyeglasses and expressions.
We also shows the control over two age directions which are young and old in~\cref{figure:Continuous_Editing_Age}.
Along the two directions of editing, the hair style and expressions are preserved in all three examples, even for earrings (middle example). 
In~\cref{figure:Continuous_Editing_Race}, we show the results of editing race along Asian and black direction.
We are able to achieve a successful editing in both two directions.

Additionally, for racial attributes, distinct hairstyles are often observed, such as short hair that is commonly associated with black individuals in~\cref{figure:Continuous_Editing_Race}.

\subsubsection{Fine-grained Control across Multi-attribute}
\label{Sec:appmulbias}
Addressing social biases across multiple attributes such as gender and age requires a comprehensive approach that ensures fairness without compromising image quality. 
Our approach enables fine-grained control over \textit{multiple} attributes \textit{simultaneously}, allowing for unbiased and consistent outputs across diverse settings.

In this section, we aim at demonstrating that our \method are able to mitigate bias with multi-attribute (\eg, gender and age) rather than only one attribute (\eg, gender).
In addition, we also illustrate the ability to control over the bias level within multi-attribute. 
Specifically, we identify the bias feature indexes for the target multi-attributes (details provided in~\cref{Sec:appmethod}). 
These features are then simultaneously intervened (\eg, male and old features). Visual results presented in~\cref{figure:Gender_Age_01} illustrate that \method effectively transforms images in the male and old directions without overlap (\ie, the male and old directions do not interfere with each other). 
This demonstrates the disentanglement achieved within the sparse semantic space, as discussed in~\cref{Sec:disentangle}.
We do not compare other baselines except for H-Distribution because in~\cite{parihar2024balancing}, they mention that Latent Editing~\cite{kwon2023diffusion} are unable to mitigate bias in case of multiple attributes.

\subsubsection{Control of Other Bias Mechanisms}
\label{Sec:appcaseotherbias}
\begin{figure}[htbp!]
\centering
\includegraphics[width=0.45\textwidth]{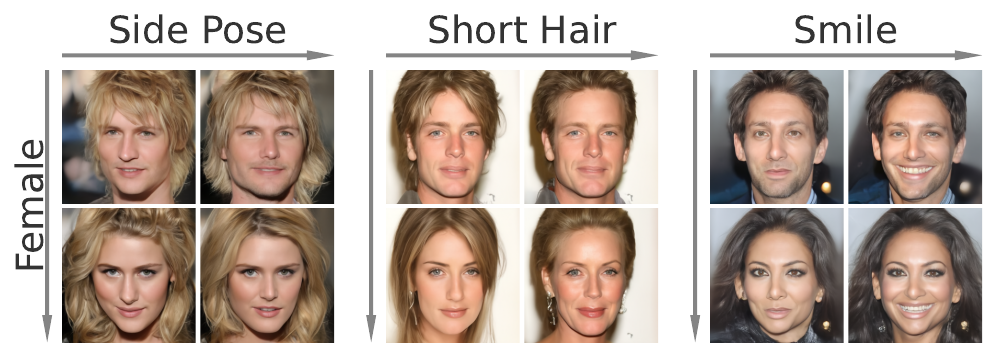}
\caption{
Decoupling and independently controlling gender and other features.
We select three features presented in~\cref{sec:bias-mech} that are ``Side Pose'', ``Short Hair'' and ``Smile''.
}
\label{fig:correlation-exp}
\end{figure}

Our \method can also disentangle gender attribute with other features.
We conduct additional experiments and present the results in~\cref{fig:correlation-exp}.
As illustrated in the figure, we can independently control gender attributes (male and female) along with features in our wide investigation in~\cref{sec:bias-mech} such as ``Side Pose'', ``Short Hair'' and ``Smile''.

\section{Further Discussions}
\subsection{Complexity of Social Bias}

Defining social bias in AI systems is inherently fraught with complexities due to the fluid, multidimensional nature of social attributes such as gender and ethnicity, which resist discrete categorization and are shaped by sociocultural contexts~\cite{Racialinmlbenthall2019}. 
Unlike measurable technical metrics, biases in systems like text-to-image models often reflecting historical inequities or stereotyping patterns rather than explicit labels. 
For instance, synthetic depictions of fictive humans lack inherent social identities, forcing evaluators to rely on subjective interpretations of visual features (\eg, skin tone or hairstyles) that may not align with real-world self-identification.

The relative instability effects for race attributes in~\cref{Tab:uncond,Tab:t2i} may possibly due to the complexity of social bias and the evaluation is inherently challenging~\cite{luccioni2024stable},  which requires more robust assessment methods.
Additionally, these attributes in dataset may be imbalanced, making model learning unstable.
However, our method balances debiasing effect (FD \cite{parihar2024balancing}) and generation quality (FID~\cite{heusel2017gans} and CLIP-score used in~\cite{shenfinetuning}).

\subsection{Monosemanticity in Sparse Autoencoder}
\label{Sec:appmonosemanticityinsae}
A monosemantic feature corresponds to one individual concept recognized by the model, in contrast to polysemantic neurons associating multiple unrelated concepts~\cite{bereska2024mechanistic}.
While an SAE automatically disentangles neuron spaces~\cite{makhzani2013k}, its 
semantic space may not fully align with human-interpretable concepts (see Sec. 3.1 of~\cite{bereska2024mechanistic}).

According to~\cite{makhzani2013k}, the used k-SAE encourages feature orthogonality, helps disentangle neuron space towards monosemanticity. 
It inherently supports disentangled feature learning~\cite{makhzani2013k}. 
It is empirically supported by low pairwise cosine similarity among the learned SAE features in \method with mean value of 0.04 and maximum value 0.17, which somehow indicates dissimilarity and disentanglement in sparse feature space.

Bias attribute, \eg, gender, is a human-defined compound concept, may relate to multiple features discovered by SAEs, for example, short-hair and mustache features. 
Our method identifies these associations and can further separate them (as shown in~\cref{fig:correlation-exp}).
Features capture aspects of our target concepts, but they may not fully represent the concepts themselves. 
To better align these features with our interpretable concepts is a valuable future direction.

\subsection{Quantitative Metric for Changes along One Direction}
Our work explicitly measures specific attribute balancing (FD~\cite{parihar2024balancing})
\begin{figure}[htbp!]
\centering
\includegraphics[width=0.4\textwidth]{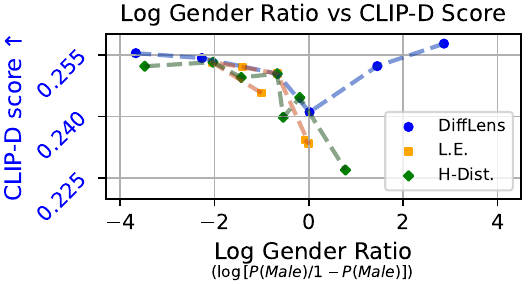}
\caption{
Alignment (CLIP-D score) to specific attribute changing across gender ratio (x-axis). 
The Log Gender Ratio reflects the log of male to female ratio in the generated images, with 0 indicating balance. 
Our \method offers better alignment with each direction. 
}
\label{fig:clipd}
\end{figure}
and overall generation quality (FID~\cite{heusel2017gans} and  CLIP-score that is used in~\cite{shenfinetuning}). 
To further assess specific attribute changes, we introduce 
\begin{equation}
   \text{CLIP-D} = \frac{\mathbf{e}_{\text{attr}} \cdot \mathbf{e}_{\text{img}}^{\text{gen}}}{\|\mathbf{e}_{\text{attr}}\| \|\mathbf{e}_{\text{img}}^{\text{gen}}\|}\;,
\end{equation}
where $\mathbf{e}_{\text{attr}}$ is the target attribute text (e.g., ``male'') embedding and $\mathbf{e}^{\text{gen}}_{\text{img}}$ is the generated image embedding. 
\Cref{fig:clipd} shows that our \method achieves better alignment to specific attribute change direction (higher CLIP-D) across varying gender bias ratios.

\subsection{Ethics Discussion}
In this work, we try to address fairness in both unconditional and conditional diffusion models by proposing a framework that identifies and isolates bias mechanisms and control bias levels in generated contents.
While our method does not prescribe a universal definition of fairness because ethical interpretations may vary across contexts. 
Our method should enable practitioners to enforce distributions considered appropriate for their applications.

All experiments were conducted using publicly available datasets and pre-trained models that are permitted for academic research. 
Our study prioritizes transparency in methodology and outcomes. 
We urge future researchers to critically evaluate the societal implications of their chosen distributions to mitigate unintended harms.

Our approach treats debiased attributes as discrete categories, thereby overlooking individuals who do not neatly fit traditional classifications (\eg, those identifying as non-binary gender or of mixed race).
This is a significant research question and needs to be addressed by future work. 
\end{document}